\documentclass[sigconf]{aamas} 

\usepackage{balance} 

\setcopyright{ifaamas}
\acmConference[AAMAS '24]{Proc.\@ of the 23rd International Conference
on Autonomous Agents and Multiagent Systems (AAMAS 2024)}{May 6 -- 10, 2024}
{Auckland, New Zealand}{N.~Alechina, V.~Dignum, M.~Dastani, J.S.~Sichman (eds.)}
\copyrightyear{2024}
\acmYear{2024}
\acmDOI{}
\acmPrice{}
\acmISBN{}

\acmSubmissionID{100}

\title[Pretrained Models for Sequential Decision-Making]{Multimodal Pretrained Models for Verifiable Sequential Decision-Making: Planning, Grounding, and Perception}

\author{Yunhao Yang}
\affiliation{
  \institution{University of Texas at Austin}
  \city{Austin, TX}
  \country{United States}}
\email{yunhaoyang234@utexas.edu}

\author{Cyrus Neary}
\affiliation{
  \institution{University of Texas at Austin}
  \city{Austin, TX}
  \country{United States}}
\email{cneary@utexas.edu}

\author{Ufuk Topcu}
\affiliation{
  \institution{University of Texas at Austin}
  \city{Austin, TX}
  \country{United States}}
\email{utopcu@utexas.edu}

\begin{abstract}

Recently developed pretrained models can encode rich world knowledge expressed in multiple modalities, such as text and images. However, the outputs of these models cannot be integrated into algorithms to solve sequential decision-making tasks. We develop an algorithm that utilizes the knowledge from pretrained models to construct and verify controllers for sequential decision-making tasks, and to ground these controllers to task environments through visual observations with formal guarantees. 
In particular, the algorithm queries a pretrained model with a user-provided, text-based task description and uses the model's output to construct an \textit{automaton-based controller} that encodes the model's task-relevant knowledge. It allows formal verification of whether the knowledge encoded in the controller is consistent with other independently available knowledge, which may include abstract information on the environment or user-provided specifications. 
Next, the algorithm leverages the vision and language capabilities of pretrained models to link the observations from the task environment to the text-based control logic from the controller (e.g., actions and conditions that trigger the actions). 
We propose a mechanism to provide probabilistic guarantees on whether the controller satisfies the user-provided specifications under perceptual uncertainties.
We demonstrate the algorithm's ability to construct, verify, and ground automaton-based controllers through a suite of real-world tasks, including daily life and robot manipulation tasks.

\end{abstract}


\newcommand{\BibTeX}{\rm B\kern-.05em{\sc i\kern-.025em b}\kern-.08em\TeX}

\makeatletter
\gdef\@copyrightpermission{
	\begin{minipage}{0.3\columnwidth}
		\href{https://creativecommons.org/licenses/by/4.0/}{\includegraphics[width=0.90\textwidth]{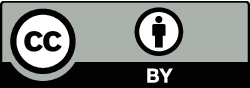}}
	\end{minipage}\hfill
	\begin{minipage}{0.7\columnwidth}
		\href{https://creativecommons.org/licenses/by/4.0/}{This work is licensed under a Creative Commons Attribution International 4.0 License.}
	\end{minipage}
	\vspace{5pt}
}
\makeatother

\usepackage{algorithm}
\usepackage{algpseudocode}
\usepackage{xcolor}
\definecolor{ForestGreen}{RGB}{34,139,34}
\usepackage{subcaption}
\usepackage{newfloat}
\usepackage{adjustbox}
\usepackage{listings}
\usepackage{tikz}
\usepackage{mathtools} 
\usepackage{amsfonts} %
\usepackage{pgfplots}
\pgfplotsset{compat=1.18}
\usetikzlibrary[pgfplots.groupplots]
\usepackage{wrapfig}

\usetikzlibrary{arrows, automata, positioning}
\tikzset{auto, >=stealth}
\tikzset{every edge/.append style={shorten >= 1pt}}
\tikzset{
    main node/.style={circle,draw,minimum size=1cm,inner sep=0pt},
}
\usetikzlibrary{arrows.meta}
\usetikzlibrary{calc}

\DeclareCaptionStyle{ruled}{labelfont=normalfont,labelsep=colon,strut=off} 
\floatstyle{ruled}
\newfloat{listing}{tb}{lst}{}
\floatname{listing}{Listing}

\usepackage{macros}
\usepackage[english]{babel}
\usepackage{amsthm}
\usepackage{graphicx, array, blindtext}
\usepackage{dblfloatfix}
\theoremstyle{definition}

\newtheorem{theorem}{Theorem}

\begin{document}


\pagestyle{fancy}
\fancyhead{}


\maketitle 


\section{Introduction}

\begin{figure}[t]
    \centering
    \includegraphics[width=\linewidth]{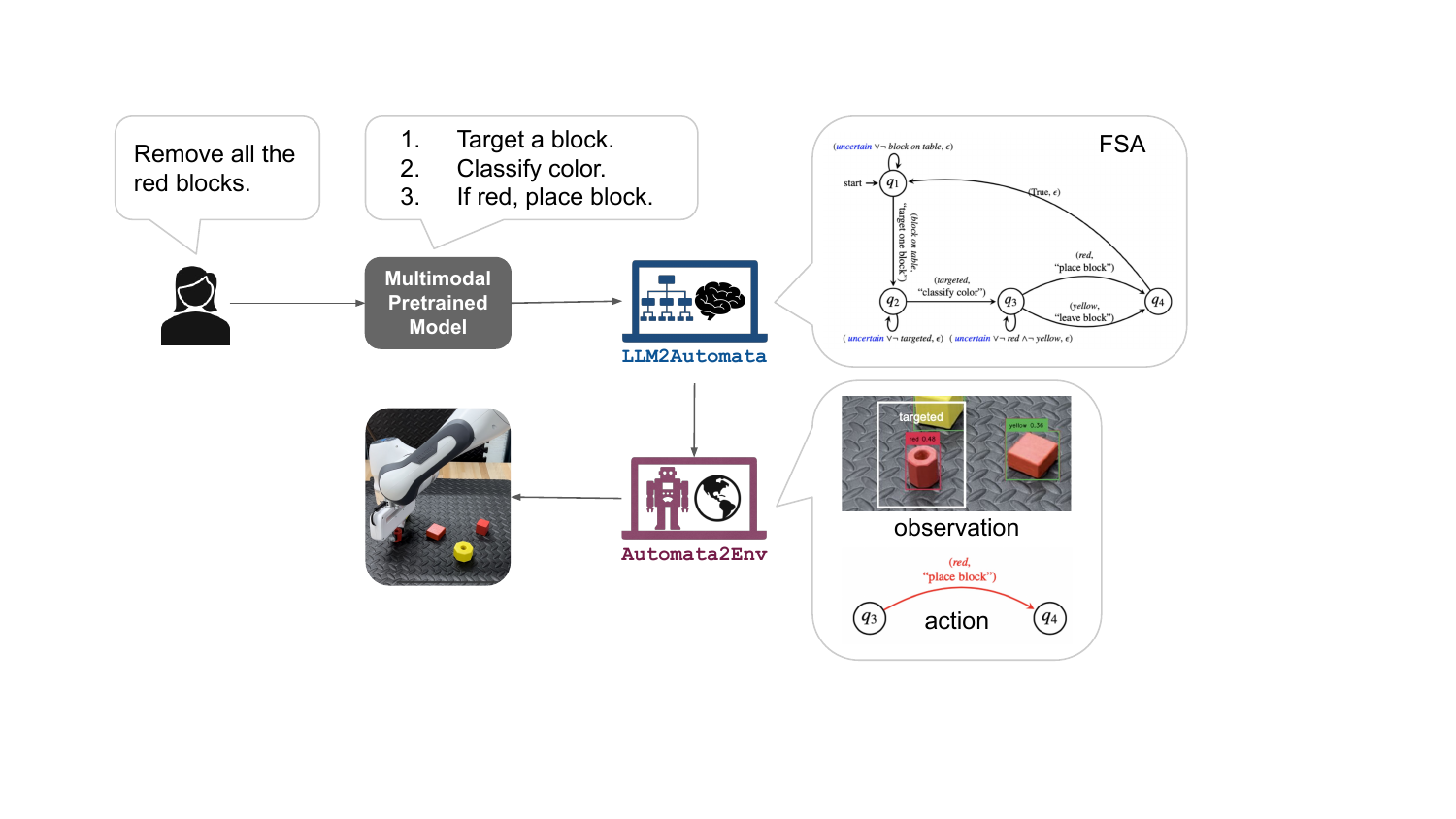}
    \caption{A real-world example that applies the proposed pipeline to a robot arm manipulation task.}
    \label{fig: example}
\end{figure}

While the rapidly emerging capabilities of multimodal pretrained models (also referred to as foundation models or base models) in question answering, code synthesis, and image generation offer new opportunities for autonomous systems, a gap exists between the text-based and image-based outputs of these models and algorithms for solving sequential decision-making tasks. 
Additional methods are required to integrate the outputs of these pretrained models into autonomous systems that can perceive and react to an environment in order to fulfill a task.
Additionally, it is hard, if not impossible, to formally verify whether autonomous systems implementing such pretrained models satisfy user-provided specifications.

Towards filling the gap between multimodal pretrained models and sequential decision-making, we develop a pipeline that integrates the outputs of pretrained models into downstream design steps, e.g., control policy synthesis or reinforcement learning, and provides a systematic way to ground the knowledge from such models. 
Specifically, we develop an algorithm to construct \emph{automaton-based controllers} representing the knowledge from the pretrained models.
Such representations can be formally verified against knowledge from other independently available sources, such as abstract information on the environment or user-provided specifications. 
This verification step ensures consistency between the knowledge encoded in the pretrained model and the knowledge from other independent sources.
To implement the controllers in their task environments, we leverage the multimodal capabilities of the pretrained models, i.e., simultaneous vision and language understanding, to ground these controllers through visual perception.

The proposed method \grounding{} links image-based observations from the task environment to the controller's text-based propositions representing the environment's conditions.
Specifically, \grounding{} collects visual observations and uses the vision and language capabilities of the employed pretrained models to evaluate the truth values of conditions from the controller. The controller then uses these truth values to select its next action.
We propose a mechanism that halts the controller's actions under perceptual uncertainties, i.e., potential misclassifications raised by the pretrained model.
By doing so, this mechanism can provide probabilistic guarantees of whether the controller satisfies user-provided specifications while operating in the task environment.
It helps to ensure the autonomous agent's safety with respect to provided mission specifications under perceptual uncertainties.
For verification purposes, we use \glspl{aut} to represent the controllers.

To construct these \gls{aut}-based controllers, we develop an algorithm named \GLMtoFSA{} that constructs a controller encoding the task knowledge obtained from the pretrained model.
The algorithm builds upon the authors' recently presented algorithm, \texttt{GLM2FSA} \cite{Yang2022AutomatonBasedRO}:
It similarly queries the pretrained model to obtain text-based task knowledge, parses the text to extract actions, and defines a set of rules (grammar) to transform these actions into an \gls{aut}.
In contrast to \texttt{GLM2FSA}, \GLMtoFSA{} explicitly queries the pretrained model for the environment conditions before and after each action is taken and encodes them into the constructed controller. 
This distinction of \GLMtoFSA{} is proposed to facilitate the grounding method \grounding{}, which connects these conditions to image-based observations of the task environment.

We demonstrate the algorithms' capabilities on sequential decision-making tasks through a variety of case studies. 
We provide proof-of-concept examples of commonsense tasks (e.g., cross the road) and real-robot tasks (e.g., robot arm manipulation).
Figure \ref{fig: example} illustrates the major components of the proposed pipeline when it is applied to a robot arm manipulation task.
These examples show the algorithms' ability to construct verifiable knowledge representations and to ground these representations in real-world environments through visual observations with perceptual uncertainties. 

\section{Related Work}

\paragraph{Formal Representations of Textual Knowledge.}
Many works have developed methods to construct symbolic representations of task knowledge from natural language descriptions. 
Several works construct knowledge graphs from textual descriptions of given tasks \citep{KnowledgeGraph, Rezaei2022UtilizingLM, He2022AcquiringAM}, or analyze causalities between the textual step descriptions and build causal graphs \citep{Lu2022NeuroSymbolicCL}.
However, the graphs resulting from these works are not directly useful in algorithms for sequential decision-making, nor are they formally verifiable.
Another work builds automaton-based representations of task-relevant knowledge from text-based descriptions of tasks \cite{Yang2022AutomatonBasedRO}.
These representations are both formally verifiable and directly applicable to algorithms for sequential decision-making.
However, in contrast with \cite{Yang2022AutomatonBasedRO}, we not only generate automaton-based representations but also ground the generated representations to the task environment through image-based perceptions.

\paragraph{Multimodal Models in Sequential Decision-Making.}
A work \citep{neural-symbolic} generates static high-level plans and matches them to the closest admissible action.
Some other works \citep{huang2022language, Lin2023Text2MotionFN, Liu2023LLMPEL, Singh2022ProgPromptGS} generate zero-shot plans for sequential decision-making tasks from querying generative language models. These works require a set of pre-defined actions, which limit their generalization capability. 
Another work \citep{vemprala2023chatgpt} uses large language models to generate executable code or API for robots.
These works lack a procedure to ensure the correctness or safety of their generated plans or executable actions. In contrast, the automaton-based representation we constructed enables others to formally verify the plans against some mission or safety specifications.

\paragraph{Multimodal Models Grounding and Perceptions.}
Several works \citep{Song2022LLMPlannerFG, brohan2023can, shah2022robotic, huang2022inner} match textual plans to image observations and perform actions based on the perceptual outputs. A work \citep{Wang2023DescribeEP} recursively generates plans based on visual observations. Other works \citep{Lu2023MultimodalPP, LinHLGS023} match texts to images by generating visual-grounded textual plans from generative models and images.

However, none of these works guarantees the safety or correctness during the grounding procedure, especially under perceptual uncertainties. They assume the vision models can correctly classify the content within the image and correctly make plans or actions accordingly. In contrast, we consider the uncertainties in the image observations from the environments and enable the capability of formally verifying the automaton-based representations against provided specifications over the task environment with uncertainties.

There are multimodal pretrained models with vision and image capabilities that can interpret the content within the images and connect images to natural language. CLIP \citep{clip} measures the text-image consistency. Many other models \cite{yolo, rcnn} can detect objects described in text from a given image. However, they have a fixed set of vocabularies to define objects. Open-vocabulary object detection models \citep{liu2023grounding, kirillov2023segmentanything, GuLKC22, LiZZYLZWYZHCG22} remove the constraints on vocabularies, which we will use for connecting the automaton-based representations to the task environment.
\begin{figure*}[t]
    \centering
    \includegraphics[width=0.8\linewidth]{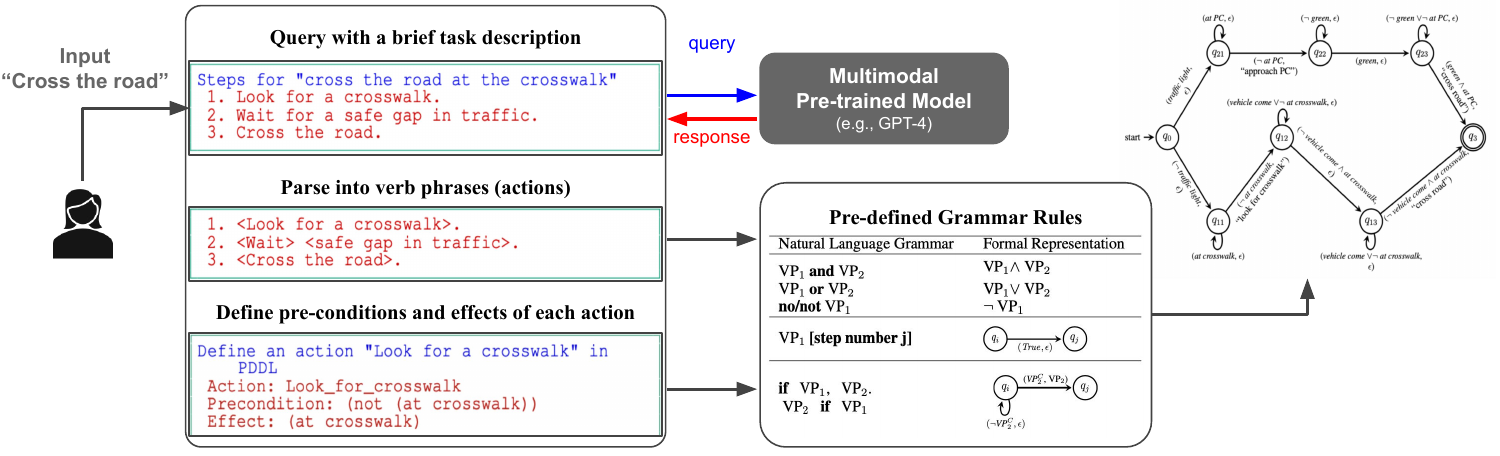}
    \caption{
    Demonstration of using the \GLMtoFSA{} algorithm to construct the controller through queries to the \gls{glm}.
    }
    \label{fig: construct}
\end{figure*}

\section{Preliminaries}

\paragraph{Multimodal Pretrained Models.}
Multimodal pretrained models (also referred to as foundation models \cite{bert} or base models \cite{clip}) are capable of processing, understanding, and generating data across multiple formats, such as images, text, and audio. 
These models are pretrained on large training datasets, and they have demonstrated strong empirical performance across a variety of tasks, such as question-answering and next-word prediction, even without further task-specific fine-tuning \cite{brown2020GPT3}.

The Generative Pretrained Transformer (GPT) series of models \citep{openai2023gpt4, brown2020GPT3} consists of the most well-known multimodal pretrained models that can generate natural language or other data formats. 
In addition to GPT, pretrained models such as PaLM \citep{Chowdhery2022PaLMSL}, BLOOM \citep{Scao2022BLOOMA1}, Codex \citep{codex}, and Megatron \citep{Smith2022UsingDA} also have the capability of generating outputs in natural language or other formats.
Language generation is the core capability of these models, which we will use in the rest of the paper. Hence we denote this category of multimodal pretrained models as \Glspl{glm}.

\emph{Vision-language models} such as CLIP \citep{clip}, Yolo \citep{yolo}, and the Segment Anything Model \citep{kirillov2023segmentanything} are another type of multimodal pretrained model. CLIP takes an image and a set of texts as inputs, and measures the image-text consistency. Yolo, R-CNN \citep{rcnn} and Segment Anything Model are object detection models, which take an image and a set of words that describe objects, and classify whether the objects appear in the image.
These models are capable of processing and understanding texts and images but are not capable of content generation.

\paragraph{Finite State Automaton.}

A \glsreset{aut}\gls{aut} is a tuple
$\Aut =
\langle
    \AutSymbsIn,
    \AutSymbsOut,
    \AutStates,
    \Autstate_0,
    \AutTransFunc,
    \AutLabelFunc
\rangle$
where
$\AutSymbsIn$ is the input alphabet (the set of input symbols),
$\AutSymbsOut$ is the output alphabet (the set of output symbols),
$\Autstate_0 \in \AutStates$ is the initial state,
$\AutTransFunc: \AutStates \times \AutSymbsIn \times \AutStates \rightarrow \{0, 1\}$ is the transition function, 
and $\AutLabelFunc: \AutStates \times \AutSymbsIn \times \AutStates \rightarrow \AutSymbsOut$ is the output function.

We use \(\AutProps\) to denote the set of atomic propositions, which we use to define the input alphabet, \(\AutSymbsIn \coloneqq 2^{\AutProps}\).
In words, any given input symbol $\Autsymbin \in \AutSymbsIn$ consists of a set of atomic propositions from \(\AutProps\) that currently evaluate to True.
A \emph{propositional logic formula} is based on one or more atomic propositions in $\AutProps$.
A transition from $\Autstate_i$ to $\Autstate_j$ exists if $\AutTransFunc(\Autstate_i, \varphi, \Autstate_j) = 1$, the current state is $\Autstate_i$, and the propositional logic formula $\varphi$ is true. 
Note that we define the \gls{aut} transitions to possibly be non-deterministic, i.e., multiple transitions are possible under the same input symbol from a given \gls{aut} state.


\paragraph{Controllers and Models.}

In this work, we refer to the automaton-based representation of task knowledge as a \emph{controller}:
a system component responsible for making decisions and taking actions based on the system's state. A controller is represented as mapping the system's current state to an action, which can be interpreted as a control input or a setpoint. Mathematically, we use an \gls{aut} $
\langle
    \AutSymbsIn,
    \AutSymbsOut,
    \AutStates,
    \Autstate_0,
    \AutTransFunc,
    \AutLabelFunc
\rangle$
to represent the controller, whose input alphabet $\AutSymbsIn$ indicates all possible observations of the environment and output alphabet $\AutSymbsOut$ indicates all possible actions. 
We additionally allow for a ``no operation'' action $\noop \in \AutSymbsOut$.

The controller's goal is to adjust the control input so that the system's state evolves in a way that satisfies externally provided requirements or properties. These requirements or properties are often specified using formal languages, such as \gls{ltl} \citep{Biggar2020AFF}.

A \emph{model} is a transition system that may represent either the dynamics of the task environment or knowledge from other independent sources. A \emph{model}
$\Aut[model] \coloneqq
\langle
    \AutSymbsIn[model],
    \AutSymbsOut[model],
    \AutStates[model],
    \AutTransFunc[model],
    \AutLabelFunc[model]
\rangle$
consists of input alphabet $\AutSymbsIn[model] \coloneqq 2^{\AutProps[model]}$ is a set of input symbols, where $\AutProps[model]$ is defined as the actions.
$\AutStates[model]$ is a finite set of states,
$\AutTransFunc[model]: \AutStates[model] \times \AutSymbsIn[model] \times \AutStates[model] \rightarrow \{0, 1\}$ is a non-deterministic transition function, and $\AutLabelFunc[model]: \AutStates[model] \to \AutSymbsOut[model]$ is a labeling function, 
where $\AutSymbsOut[model] = 2^{\overline{P}}$ and $\overline{P}$ is a set of atomic propositions representing conditions of the environment.

\paragraph{The Planning Domain Definition Language.}
A \gls{pddl} \citep{pddl} is a formal language used in artificial intelligence and automated planning to define a planning problem. We use \gls{pddl} to describe the possible initial states of a problem, the desired goal, and the actions that can be taken to transform the initial state into the goal state.
\gls{pddl} provides a standardized syntax for specifying a set of predicates---atomic propositions---describing the states of the task, the actions, and the goal specification.

Each action $\Autsymbout$ in \gls{pddl} has a name, a \emph{precondition} that must be satisfied before the action can be performed, and an of \emph{effect} that describes how the state of the environment will change after the action is performed. The preconditions and effects are expressed as sets of atomic propositions.

\section{Task Controller Construction}

We develop an algorithm, \GLMtoFSA{}, that takes a brief task description in textual form from the task designer and returns an \gls{aut} representing the task controller that can be verified against the specifications given by the task designer. 

The algorithm \GLMtoFSA{} takes a brief text description of a task and constructs an FSA to represent the controller of the given task. Specifically, the algorithm sends the text description as the input prompt (in blue) to a \gls{glm} and obtains the \gls{glm}'s response (in red), which is a list of steps for achieving the task in textual form:
\noindent\begin{minipage}{\linewidth}
\begin{lstlisting}[language=completion]
    <prompt>Steps for <emph>task description</emph></prompt><completion> 
    step_number_1. <emph>step description</emph>
    step_number_2. <emph>step description</emph>
    ...</completion>
\end{lstlisting}
\end{minipage}

The algorithm uses the semantic parsing method introduced in GLM2FSA \citep{Yang2022AutomatonBasedRO} to parse each step description into \emph{verb phrases (VP)} and connective keywords. A list of pre-defined keywords is provided in Table \ref{tab: grammar}. A verb phrase consists of a verb and its noun dependencies. 
Each step corresponds to a state in the FSA. 
Meanwhile, each verb phrase VP in the step description represents an \emph{action}, and the algorithm queries the \gls{glm} to extract the precondition and effect of this action in the form of \gls{pddl}:
\begin{lstlisting}[language=completion]
    <prompt>Define an action <emph>"action name"</emph> in PDDL</prompt>
    <completion> Action: <emph>action name</emph></completion>
    <completion> Precondition: <emph>a set of propositions</emph></completion>
    <completion> Effect: <emph>a set of propositions</emph></completion>
\end{lstlisting}
We use the extracted verb phrase VP$_i$ to define the action name, and we use VP$_i^C$ and VP$_i^E$ to denote the precondition and effect of the action, respectively. Then, the algorithm follows the rules illustrated in Table \ref{tab: grammar} to transform natural language into propositions or automaton transitions. Each step description is translated into a state in the FSA and a set of outgoing transitions from this state.

We note that in contrast to the GLM2FSA algorithm presented in \cite{Yang2022AutomatonBasedRO}, we query the \gls{glm} for the preconditions and effects of each action and encode them into the constructed controller.
These preconditions and effects are descriptions of the task environment prior to and after taking some actions. 
This explicit representation of the actions' preconditions and effects is required for the methodology we propose to ground the constructed automaton-based controllers to their task environments via image-based observations, described in Section \ref{sec:grounding_and_perception}.

\begin{table}[t]
\centering
\begin{tabular}{m{0.25\linewidth} m{0.35\linewidth} m{0.3\linewidth}}
\hline
Grammar & Formal Representation & Example\\
\hline
\vspace{0.2cm}
VP$_1$ \textbf{and} VP$_2$ & VP$_1 \land$ VP$_2$ & [green light] [and] [no car] \\ 
VP$_1$ \textbf{or} VP$_2$ & VP$_1 \vee$ VP$_2$ & [traffic light] [or] [crosswalk] \\
\textbf{no/not} VP$_1$ & $\neg$ VP$_1$ & [no] [car] \\
\hline

\shortstack{VP$_1$ \textbf{[step j]}} & \vspace{0.2cm}\begin{tikzpicture}[thick,scale=.6, node distance=2.2cm, every node/.style={transform shape}]
	\node[state] (0) at (0, 0) {\Large $q_i$};
	\node[state] (2) at (3, 0) {\Large $q_j$};

	\draw[->, shorten >=1pt] (0) to[left] node[below, align=center] {$(\ltrue,\noop)$} (2);
\end{tikzpicture} & [go to step]  [1]\\
\hline

\shortstack{ \textbf{if} $\:$ VP$_1$, $\:$ VP$_2$. \\ VP$_2$ $\:$ \textbf{if} $\:$ VP$_1$} &
\vspace{0.2cm}\begin{tikzpicture}[thick,scale=.6, node distance=2.2cm, every node/.style={transform shape}]
	\node[state] (0) at (0, 0) {\Large $q_i$};
	\node[state] (1) at (3, 0) {\Large $q_{j}$};

	\draw[->, shorten >=1pt] (0) to[left] node[above, align=center, sloped] {(\textit{VP$_2^C$}, VP$_2$)} (1);
    \draw[->, shorten >=1pt] (0) to[loop below] node[align=center] {$(\lnot\textit{VP}_2^C, \noop)$} ();
\end{tikzpicture} & [if] [green light], [cross] \\
\hline

\shortstack{ \textbf{wait} VP$_1$ VP$_2$ \\ VP$_2$ \textbf{after} VP$_1$ } &
\vspace{0.2cm} \begin{tikzpicture}[thick,scale=.6, node distance=2.2cm, every node/.style={transform shape}]
	\node[state] (0) at (0, 0) {\Large $q_i$};
	\node[state] (1) at (3, 0) {\Large $q_{i+1}$};

	\draw[->, shorten >=1pt] (0) to[left] node[below, align=center] {$(\textit{VP}_1^E, \text{VP}_2)$} (1);
    \draw[->, shorten >=1pt] (0) to[loop below] node[align=center] {$(\lnot\textit{VP}_1^E, \noop)$} ();
\end{tikzpicture} & [wait] [green light] [cross]\\

\shortstack{ VP$_2$ \textbf{until} VP$_1$} &
 \begin{tikzpicture}[thick,scale=.6, node distance=2.2cm, every node/.style={transform shape}]
	\node[state] (0) at (0, 0) {\Large $q_i$};
	\node[state] (1) at (3, 0) {\Large $q_{i+1}$};

	\draw[->, shorten >=1pt] (0) to[left] node[below, align=center] {$(\textit{VP}_1, \noop)$} (1);
    \draw[->, shorten >=1pt] (0) to[loop below] node[align=center] {$(\lnot\textit{VP}_1, \text{VP}_2)$} ();
\end{tikzpicture} & [not cross] [until] [green light]\\
\hline

\shortstack{VP$_1$} &
\vspace{0.2cm} \begin{tikzpicture}[thick,scale=.6, node distance=2.2cm, every node/.style={transform shape}]
	\node[state] (0) at (0, 0) {\Large $q_i$};
	\node[state] (1) at (3, 0) {\Large $q_{i+1}$};

	\draw[->, shorten >=1pt] (0) to[left] node[below, align=center] {$(\textit{VP}_1^C, \text{VP}_1)$} (1);
    \draw[->, shorten >=1pt] (0) to[loop left] node[below, align=center] {\small $(\neg \textit{VP}_1^C, \epsilon)$} ();
\end{tikzpicture} & [cross road]\\
\hline

\end{tabular}
\caption{
    Rules to convert natural language grammar to formal representations (propositions or FSA transitions). The keywords that define the grammar are in bold.
}
\label{tab: grammar}
\end{table}

\subsection{Verifying against External Knowledge}
\label{sec:model_extraction}
An automaton-based model encodes the dynamics of the task environment or the task-relevant knowledge from external knowledge sources. Users can provide automaton-based models to verify whether the knowledge from the \gls{glm} is consistent with the user-provided knowledge or requirements.

Once we have the controller and the model, we use the model to formally verify whether the controller satisfies user-provided specifications.
In the verification procedure, we build a product automaton $\Aut[product] = \Aut[model] \otimes \Aut[controller]$ describing the interactions of the controller $\Aut[controller]$ with the model $\Aut[model]$. A \emph{product automaton} is an \gls{aut}
$\Aut[product] =
\Aut[model] \otimes \Aut[controller] \coloneqq
\langle
    \AutStates[product],
    \AutTransFunc[product],
    \Autstate[product-init],
    \AutLabelFunc[product]
\rangle
$
as follows:
\begin{align*}
\AutStates[product] &\coloneqq \AutStates[model] \times \AutStates
\\
\AutTransFunc[product]( (p, q))
&\coloneqq
\left\{
    (p', q') \in \AutStates[product] \middle|
    \AutTransFunc(q, c, q') = 1
    \land
    \AutTransFunc[model](p, \Autsymbout, p') = 1
\right\}
\\
&\text{where } \Autsymbout = \AutLabelFunc(q, \Autsymbin, q') \text{ and } c = \AutLabelFunc[model](p)
\\
\Autstate[product-init] &\coloneqq (p, \Autstate[init]) \quad \text{where $p$ can be any state in }\Aut[model]
\\
\AutLabelFunc[product]((p, q)) &\coloneqq \AutLabelFunc[model](p) \cup \AutLabelFunc(q, \AutLabelFunc[model](p), q') 
\\
& \text{where } q' \in \AutStates \text{ and } \AutTransFunc(q, \AutLabelFunc[model](p), q') = 1.
\end{align*}

Then, we obtain a \emph{specification} $\Phi$ expressed in \emph{linear temporal logic} from the task designer or whoever wants to verify the controller. We run a model checker (e.g., NuSMV~\citep{Cimatti2002NuSMV}) to verify if the product automaton satisfies the specification,
\begin{equation}
    \label{eq: model-checking}
    \Aut[model] \otimes \Aut[controller] \models \Phi.
\end{equation}
We verify the product automaton against the specification for all the possible initial states. If the verification fails, the model checker returns a counter-example, which is a sequence of states of the product automaton $(p_1, q_1), (p_2, q_2),...$ where $p_i \in \AutStates[model], q_i \in \AutStates$.

\section{Verifiable Grounding}
\label{sec:grounding_and_perception}

\begin{figure}
    \centering
    \includegraphics[width=\linewidth]{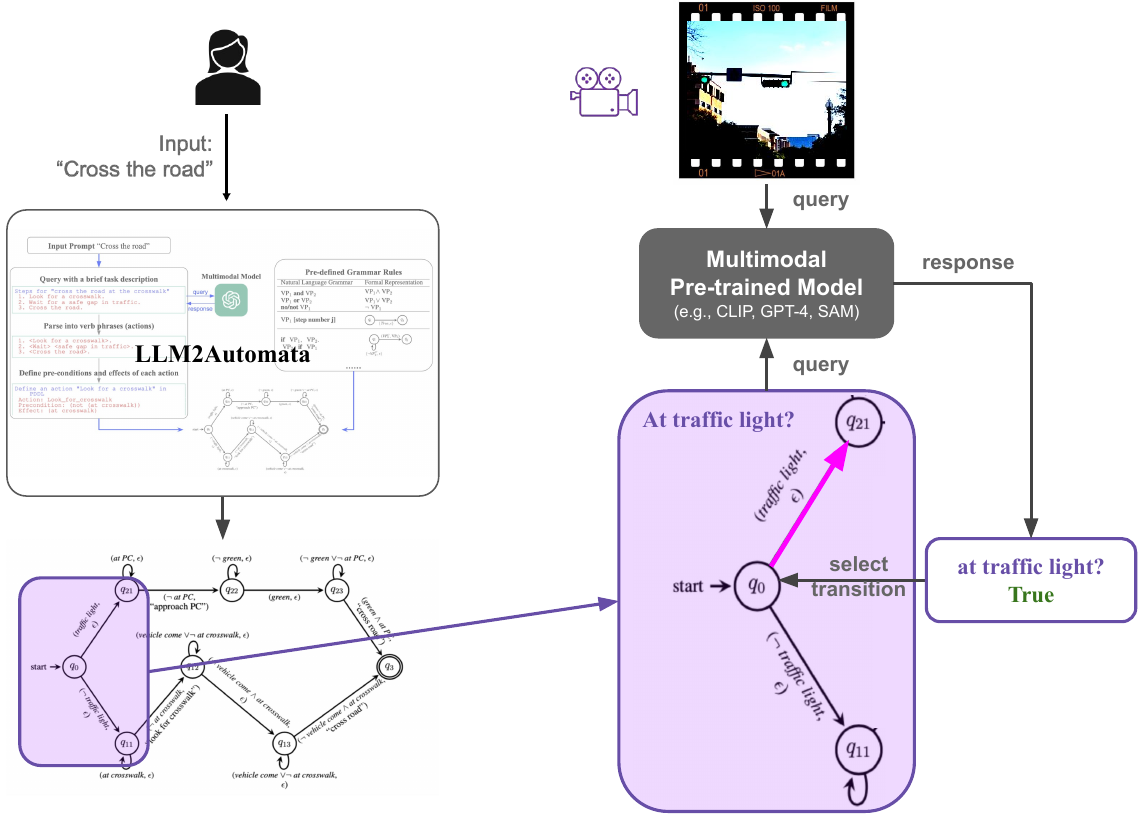}
    \caption{Demonstration of grounding the \gls{aut}-based controller to the real-world task environment through visual perceptions.}
    \label{fig: grounding}
\end{figure}

\grounding{} takes visual observations from the task environment and uses a vision-language model to determine the truth values of the atomic propositions that are relevant to the conditions specified in the controller. \grounding{} enables formal verification during the procedure of grounding the controller to the task environment.

\subsection{The Pipeline of \grounding{}}
To operate in the task environment, an agent starts from the initial state of the controller. The agent collects all the propositions $P$ from the controller and gets an image observation from the task environment. It then feeds the image and all the propositions as text into a vision-language model.
\grounding{} requires vision-language models that can output normalized scores indicating how each proposition matches the image (e.g., CLIP \citep{clip}). We refer to such scores as \emph{confidence scores}, which are commonly provided as outputs of vision-language models. A higher score means the vision-language model is more confident that the context of the proposition is within the content of the image. We incorporate these confidence scores to approximate the perceptual uncertainties.

The overall pipeline of \grounding{} is as follows: 

\paragraph{Modifying the Controller to Handle Uncertainties}
We first add \UNC{} as an additional atomic proposition and modify the controller by adding a self-transition $\AutTransFunc (q_i, \UNC{}, q_i) = 1$ to each state $q_i$. Intuitively, the controller will stay in the current state, and it will not perform any action if it gets uncertain observations. An example is presented in Figure \ref{fig: cross}.

\paragraph{Evaluating Atomic Propositions}
Second, we propose an algorithm to evaluate the truth values of propositions in image observations. 
The algorithm takes an atomic proposition in textual form, a vision-language model that can return confidence scores and numerical thresholds as inputs. 
Recall that an input symbol is a set of atomic propositions.
As opposed to ordinary binary evaluation, the algorithm evaluates an atomic proposition and assigns one of the three values: true, false, and uncertain. Algorithm \ref{alg: eval-prop} shows how we evaluate the propositions using the confidence scores from the vision-language model.

\paragraph{Taking Actions}
Third, after evaluating the set of atomic propositions, the agent chooses one transition whose input symbol (which itself is a logical formula over the atomic propositions) evaluates to true and takes corresponding actions.
A demonstration of this pipeline is in Figure \ref{fig: example}.

\begin{algorithm}[t]
  \caption{Proposition Evaluation under Uncertainty}\label{alg: eval-prop}
  \begin{algorithmic}[1]
    \Procedure{\textbf{EvalProp}}{Atomic Proposition $p$, Observation $I$, Vision-Language Model \textbf{VM}, true threshold $t$, false threshold $f$} \Comment{$p$ is a string and $I$ is an image}
    \State score = \textbf{VM}($p$, $I$)
    \If{score $\ge t$}
        \textbf{return} {\color{ForestGreen} $p$ ($p =$ true)}
    \EndIf
    \If{score $\le f$}
        \textbf{return} {\color{red} $\neg p$ ($p=$ false)}
    \EndIf
    \State \textbf{return} \UNC{}
    \EndProcedure
  \end{algorithmic}
\end{algorithm}

\subsection{Determining True and False Thresholds}
\paragraph{Selecting the Vision-Language Model}
We use the current state-of-the-art vision-language model called \gls{vlm} \citep{kirillov2023segmentanything, liu2023grounding} to evaluate the propositions from image observations. The \gls{vlm} is an open-domain object detection model, which can take any text as input and determine whether the object or scene described in the text appears in the image. The \gls{vlm} returns a confidence score for each detected object, and the score will be zero if it does not find the object in the image.


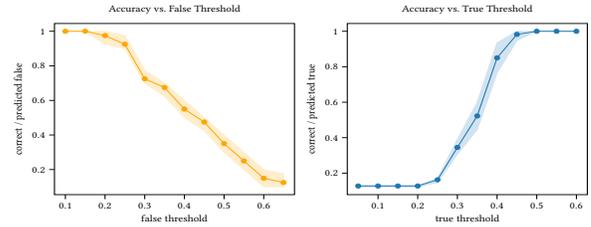
\begin{figure}
    \centering
    \resizebox{0.45\linewidth}{0.35\linewidth}{
\begin{tikzpicture}

\definecolor{darkgray176}{RGB}{176,176,176}
\definecolor{steelblue31119180}{RGB}{255,165,0}

\begin{axis}[
tick align=outside,
tick pos=left,
title={Accuracy vs. False Threshold},
x grid style={darkgray176},
xlabel={false threshold},
xmin=0.0725, xmax=0.6775,
xtick style={color=black},
y grid style={darkgray176},
ylabel={correct / predicted false},
ymin=0.0549999999999996, ymax=1.045,
ytick style={color=black}
]
\path [draw=steelblue31119180, fill=steelblue31119180, opacity=0.2]
(axis cs:0.1,1)
--(axis cs:0.1,1)
--(axis cs:0.15,1)
--(axis cs:0.2,0.925)
--(axis cs:0.25,0.9)
--(axis cs:0.3,0.699999999999999)
--(axis cs:0.35,0.625)
--(axis cs:0.4,0.5)
--(axis cs:0.45,0.424999999999999)
--(axis cs:0.5,0.300000000000001)
--(axis cs:0.55,0.200000000000001)
--(axis cs:0.6,0.0999999999999996)
--(axis cs:0.65,0.0999999999999996)
--(axis cs:0.65,0.175000000000001)
--(axis cs:0.65,0.175000000000001)
--(axis cs:0.6,0.200000000000001)
--(axis cs:0.55,0.300000000000001)
--(axis cs:0.5,0.399999999999999)
--(axis cs:0.45,0.5)
--(axis cs:0.4,0.6)
--(axis cs:0.35,0.699999999999999)
--(axis cs:0.3,0.775)
--(axis cs:0.25,0.975)
--(axis cs:0.2,1)
--(axis cs:0.15,1)
--(axis cs:0.1,1)
--cycle;

\addplot [semithick, steelblue31119180, mark=*]
table {%
0.1 1
0.15 1
0.2 0.975
0.25 0.925
0.3 0.725
0.35 0.674999999999999
0.4 0.55
0.45 0.475
0.5 0.35
0.55 0.250000000000001
0.6 0.15
0.65 0.125
};
\end{axis}

\end{tikzpicture}}
    \resizebox{0.45\linewidth}{0.35\linewidth}{
\begin{tikzpicture}

\definecolor{darkgray176}{RGB}{176,176,176}
\definecolor{steelblue31119180}{RGB}{31,119,180}

\begin{axis}[
tick align=outside,
tick pos=left,
title={Accuracy vs. True Threshold},
x grid style={darkgray176},
xlabel={true threshold},
xmin=0.0225, xmax=0.6275,
xtick style={color=black},
y grid style={darkgray176},
ylabel={correct / predicted true},
ymin=0.078625, ymax=1.043875,
ytick style={color=black}
]
\path [draw=steelblue31119180, fill=steelblue31119180, opacity=0.2]
(axis cs:0.05,0.13)
--(axis cs:0.05,0.1225)
--(axis cs:0.1,0.1225)
--(axis cs:0.15,0.1225)
--(axis cs:0.2,0.1225)
--(axis cs:0.25,0.1475)
--(axis cs:0.3,0.3075)
--(axis cs:0.35,0.445)
--(axis cs:0.4,0.765)
--(axis cs:0.45,0.9475)
--(axis cs:0.5,1)
--(axis cs:0.55,1)
--(axis cs:0.6,1)
--(axis cs:0.6,1)
--(axis cs:0.6,1)
--(axis cs:0.55,1)
--(axis cs:0.5,1)
--(axis cs:0.45,1)
--(axis cs:0.4,0.935)
--(axis cs:0.35,0.6)
--(axis cs:0.3,0.39)
--(axis cs:0.25,0.17)
--(axis cs:0.2,0.13)
--(axis cs:0.15,0.13)
--(axis cs:0.1,0.13)
--(axis cs:0.05,0.13)
--cycle;

\addplot [semithick, steelblue31119180, mark=*]
table {%
0.05 0.1275
0.1 0.1275
0.15 0.1275
0.2 0.1275
0.25 0.1625
0.3 0.345
0.35 0.5225
0.4 0.85
0.45 0.9825
0.5 1
0.55 1
0.6 1
};
\end{axis}

\end{tikzpicture}}
    \caption{The left and right figures show the proposition evaluation accuracies under different false and true thresholds.}
    \label{fig: stats}
\end{figure}

\paragraph{Validating the Vision-Language Model}
Once the vision-language model is selected, we validate the selected model on an externally provided dataset to determine the values of the true and false thresholds to be used in Algorithm \ref{alg: eval-prop}. The validation procedure is based on the following assumption:

\textbf{Assumption 0:} Validation images are drawn from the same distribution as the images from task environments. Hence the performance of a vision-language model on validation images and task images is consistent.

Under Assumption 0, we select an image dataset, Argoverse \cite{argoverse}, that contains driving scenes. We use the \gls{vlm} to detect driving-relevant objects (e.g., crosswalks, traffic lights, cars) and check whether the detection results are correct. 
We collect the confidence scores of the detection results from \gls{vlm} and the corresponding ground truth labels from the dataset.

\paragraph{Determining Thresholds}
We plot figures on false/true thresholds vs. proposition evaluation accuracy and present them in Figure \ref{fig: stats}. For each true threshold $t$, we evaluate all the detection results whose confidence scores are greater than $t$ to true and compute the percentage of ``the number of results correctly evaluated to true divided by the number of results that were evaluated to true." We denote this percentage as \emph{proposition evaluation accuracy}. Similarly, we compute the percentage of the number of results that are correctly predicted to be false, divided by the total number of results evaluated to be false.
We expect the proposition evaluation accuracies for both true and false to equal 1, which means everything that is evaluated to be true or false is correct. We denote this scenario as another assumption:

\textbf{Assumption 1 ($\mathcal{A}_1$):} If a proposition is NOT evaluated to \UNC{}, then the proposition evaluation is correct.

Now we can determine a true threshold $t$ and a false threshold $f$ according to the empirical accuracies plotted in Figure \ref{fig: stats}. 
Note that regardless of what thresholds we choose, this empirical estimate of the proposition evaluation accuracy is not 1. Hence we can obtain a probability of Assumption 1 being held ($\mathbb{P}[\mathcal{A}_1 = true]$) through Theorem \ref{thm: prob}.
\begin{theorem}
    \label{thm: prob}
    Let $\mathbf{p}_t$ be the proposition evaluation accuracy for the true threshold $t$ and $\mathbf{p}_f$ be the accuracy for the false threshold $f$. Then, for each proposition evaluation whose result is not \UNC{}, 
    \begin{equation}
        \mathbb{P}[\mathcal{A}_1=true] \ge \min (\mathbf{p}_t, \mathbf{p}_f).
    \end{equation}
\end{theorem}

\subsection{Verification}
We now have the modified controller $\overline{\mathcal{C}}$ that takes uncertainties into consideration and the selected thresholds. Given a model $\mathcal{M}$ and specifications $\Phi$, we can apply model checking to verify whether the controller, when implemented in the model, satisfies the specifications during the grounding procedure. However, we need to additionally consider the perceptual uncertainties under the selected thresholds. Instead of Equation \ref{eq: model-checking}, we apply the model checker to verify
\begin{equation}
    \label{eq: verify-ground}
    \Aut[model] \otimes \overline{\mathcal{C}} \models (\mathcal{A}_1 \implies \Phi).
\end{equation}
This model-checking procedure ensures the controller satisfies the specifications, given that we captured all the perceptual uncertainties (i.e., Assumption 1 holds). If Equation \ref{eq: verify-ground} passes the model-checking procedure, the probability of $\Phi$ being satisfied is purely based on the degree of perceptual uncertainties, which is the probability of Assumption 1 holds. Hence we can derive a new theorem.
\begin{theorem}
    \label{thm: prob2}
    Let event $\tilde e = \Aut[model] \otimes \overline{\mathcal{C}} \models (\mathcal{A}_1 \implies \Phi)$, 
    let $N_{max}$ be a user-specified parameter representing the maximum number of proposition evaluations that the controller makes during a single run.
    If $\tilde e$ is true, then the following inequality holds
    \begin{center}
        $\mathbb{P}[\Aut[model] \otimes \overline{\mathcal{C}} \models \Phi] \ge \mathbb{P}[\tilde e] \cdot \mathbb{P}[\mathcal{A}_1 = true]^{N_{max}} \ge \min \left( \mathbf{p}_t, \mathbf{p}_f \right)^{N_{max}}$.
    \end{center}
\end{theorem}
Note that we can only approximate $\mathbf{p}_t \text{ and } \mathbf{p}_f$ through empirical analysis (e.g., Figure \ref{fig: stats}). We denote the approximations of $\mathbf{p}_t \text{ and } \mathbf{p}_f$ as $\overline{p}_t \text{ and } \overline{p}_f$.
\begin{figure*}[t]
    \centering
    \includegraphics[width=0.9\linewidth]{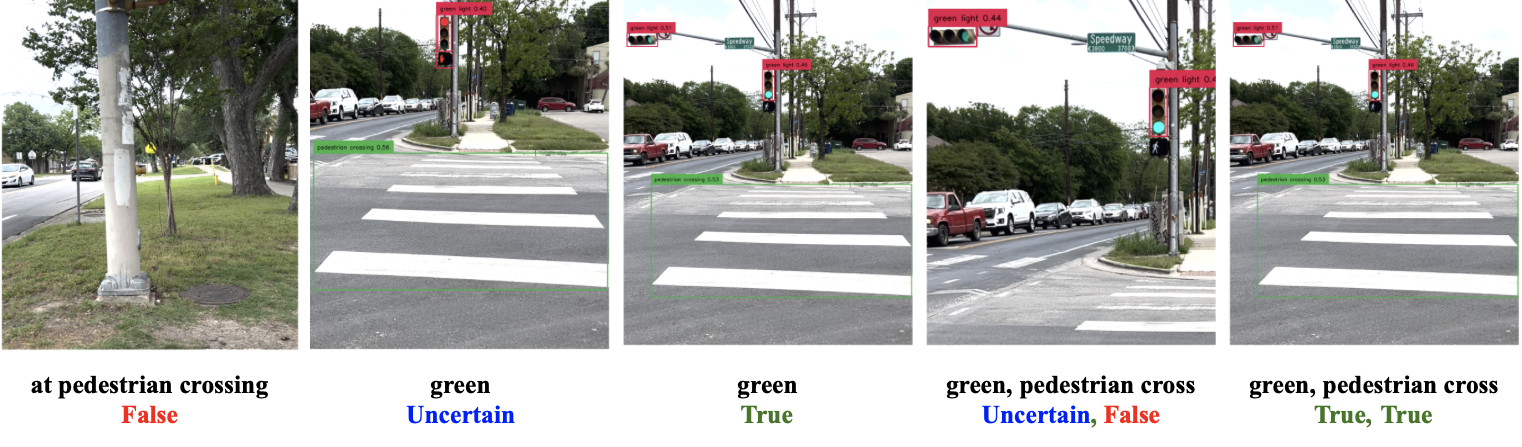}
    \caption{ A demonstration \grounding{} implementing control logic under perceptual uncertainties.
    The figure shows a sequence of observations from the real-world environment, where red and green boxes with confidence scores above are the object detection results from the \gls{vlm}.
    We use the \gls{vlm} to measure the confidence of image content and evaluate the propositions from the controller. The resulting controller's state transitions are $q_1 \rightarrow q_2 \rightarrow q_2 \rightarrow q_3 \rightarrow q_3 \rightarrow q_4$.}
    \label{fig: cross-uncertain}
\end{figure*}

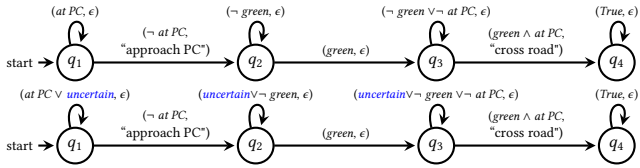
\begin{figure}[t]
    \centering
    \begin{tikzpicture}[thick,scale=.6, node distance=2.2cm, every node/.style={transform shape}]
    \node[state, initial] (21) at (0, 0) {\Large $q_{1}$};
	\node[state] (22) at (4, 0) {\Large $q_{2}$};
    \node[state] (23) at (8, 0) {\Large $q_{3}$};
	\node[state] (4) at (12, 0) {\Large $q_4$};

    \draw[->, shorten >=1pt] (21) to[loop above] node[align=center] {\small (\textit{at PC}, $\noop$)} ();
	\draw[->, shorten >=1pt, color=black] (21) to[left] node[above, align=center, sloped, color=black] {\small ($\neg$ \textit{at PC}, \\ ``approach PC")} (22);
 
    \draw[->, shorten >=1pt, color=black] (22) to[loop above] node[align=center, color=black] {\small (\textit{$\neg$ green}, $\noop$)} ();
    \draw[->, shorten >=1pt, color=black] (22) to[left] node[above, align=center, sloped, color=black] {\small (\textit{green}, $\noop$)} (23);
    
    \draw[->, shorten >=1pt] (23) to[loop above] node[align=center] {\small (\textit{$\neg$ green $\vee \neg$ at PC}, $\noop$)} ();
    \draw[->, shorten >=1pt, color=black] (23) to[left] node[above, align=center, sloped, color=black] {\small (\textit{green $\land$ at PC}, \\ ``cross road")} (4);

     \draw[->, shorten >=1pt] (4) to[loop above] node[align=center] {\small (\textit{True}, $\noop$)} ();
	
\end{tikzpicture}
    \begin{tikzpicture}[thick,scale=.6, node distance=2.2cm, every node/.style={transform shape}]
    \node[state, initial] (21) at (0, 3) {\Large $q_{1}$};
	\node[state] (22) at (4, 3) {\Large $q_{2}$};
    \node[state] (23) at (8, 3) {\Large $q_{3}$};
	\node[state] (3) at (12, 3) {\Large $q_4$};

    \draw[->, shorten >=1pt] (21) to[loop above] node[align=center] {\small (\textit{at PC $\vee$ \UNC}, $\noop$)} ();
	\draw[->, shorten >=1pt ] (21) to[left] node[above, align=center, sloped] {\small ($\neg$ \textit{at PC}, \\ ``approach PC")} (22);
 
    \draw[->, shorten >=1pt ] (22) to[loop above] node[align=center] {\small (\textit{\UNC $\vee \neg$ green},  $\noop$)} ();
    \draw[->, shorten >=1pt ] (22) to[left] node[above, align=center, sloped] {\small (\textit{green}, $\noop$)} (23);
    
    \draw[->, shorten >=1pt ] (23) to[loop above] node[align=center] {\small (\textit{\UNC $\vee \neg$ green $\vee \neg$ at PC}, $\noop$)} ();
    \draw[->, shorten >=1pt ] (23) to[left] node[above, align=center, sloped] {\small (\textit{green $\land$ at PC}, \\ ``cross road")} (3);

    \draw[->, shorten >=1pt] (3) to[loop above] node[align=center] {\small (\textit{True}, $\noop$)} ();
	
\end{tikzpicture}
    \caption{Controllers for task ``crossing the road.'' The top figure shows the controller constructed from the algorithm \GLMtoFSA{}, and the bottom figure shows the modified controller for grounding purposes. ``PC" stands for the proposition ``at pedestrian crossing" and ``green" stands for the proposition ``traffic light is green."}
    \label{fig: cross}
\end{figure}

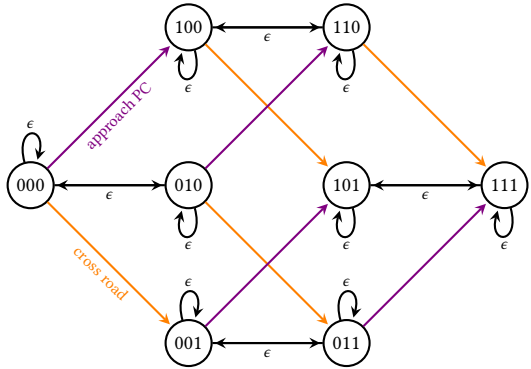
\begin{figure}[t]
    \centering
    \begin{tikzpicture}[
    scale=.7,
    node distance=2.2cm,
    thick,
    every node/.append style={transform shape},
]

\node[state] (q0)
    at (0,0)
    {\Large $000$};
\node[state] (q1)
    at (3,3)
    {\Large $100$};
\node[state] (q2)
    at (3,0)
    {\Large $010$};
\node[state] (q3)
    at (3,-3)
    {\Large $001$};
\node[state] (q4)
    at (6,3)
    {\Large $110$};
\node[state] (q5)
    at (6,0)
    {\Large $101$};
\node[state] (q6)
    at (6,-3)
    {\Large $011$};
\node[state] (q7)
    at (9,0)
    {\Large $111$};

\path[->,sloped]

(q0) 
edge[loop above] node[sloped=false]
    {$\noop$}
    ()
edge[color=violet] node[below, color=violet]
    { approach PC }
    (q1)
edge[] node[below]
    { $\noop$ }
    (q2)
edge[color=orange] node[below, color=orange]
    { cross road }
    (q3)

(q1) 
edge[loop below] node[sloped=false]
    {$\noop$}
    ()
edge[] node[below]
    { $\noop$ }
    (q4)
edge[color=orange] node[below]
    {  }
    (q5)
    
(q2) 
edge[loop below] node[sloped=false]
    {$\noop$}
    ()
edge[color=violet] node[below, color=violet]
    {  }
    (q4)
edge[color=orange] node[below]
    {  }
    (q6)
edge[] node[below]
    {  }
    (q0)

(q3) 
edge[loop above] node[sloped=false]
    {$\noop$}
    ()
edge[color=violet] node[below]
    {  }
    (q5)
edge[] node[below]
    { $\noop$ }
    (q6)

(q4) 
edge[loop below] node[sloped=false]
    {$\noop$}
    ()
edge[color=orange] node[below]
    {  }
    (q7)
edge[] node[below]
    {  }
    (q1)

(q5) 
edge[loop below] node[sloped=false]
    {$\noop$}
    ()
edge[] node[below]
    { $\noop$ }
    (q7)

(q6) 
edge[loop above] node[sloped=false]
    {$\noop$}
    ()
edge[color=violet] node[below]
    {  }
    (q7)
edge[] node[below]
    {  }
    (q3)

(q7) 
edge[loop below] node[sloped=false]
    {$\noop$}
    ()
edge[] node[below]
    {  }
    (q5)
;

\end{tikzpicture}
    \caption{Transition system that represents the environment of the ``crossing road" task. Transitions in {\color{violet} violet} and {\color{orange} orange} represent the transitions with actions ``approach traffic light" and ``cross-road," respectively. The label on the state indicates the value of propositions (``at PC," ``green," ``at other side"). For instance, 000 indicates $\neg \textit{at PC} \land \neg \textit{green} \land \neg \textit{at other side}$ and 010 indicates $\neg \textit{at PC} \land \textit{green} \land \neg \textit{at other side}$.}
    \label{fig: cross-tl-model}
\end{figure}

\begin{figure*}[t]
    \centering
    \includegraphics[height=0.3\linewidth]{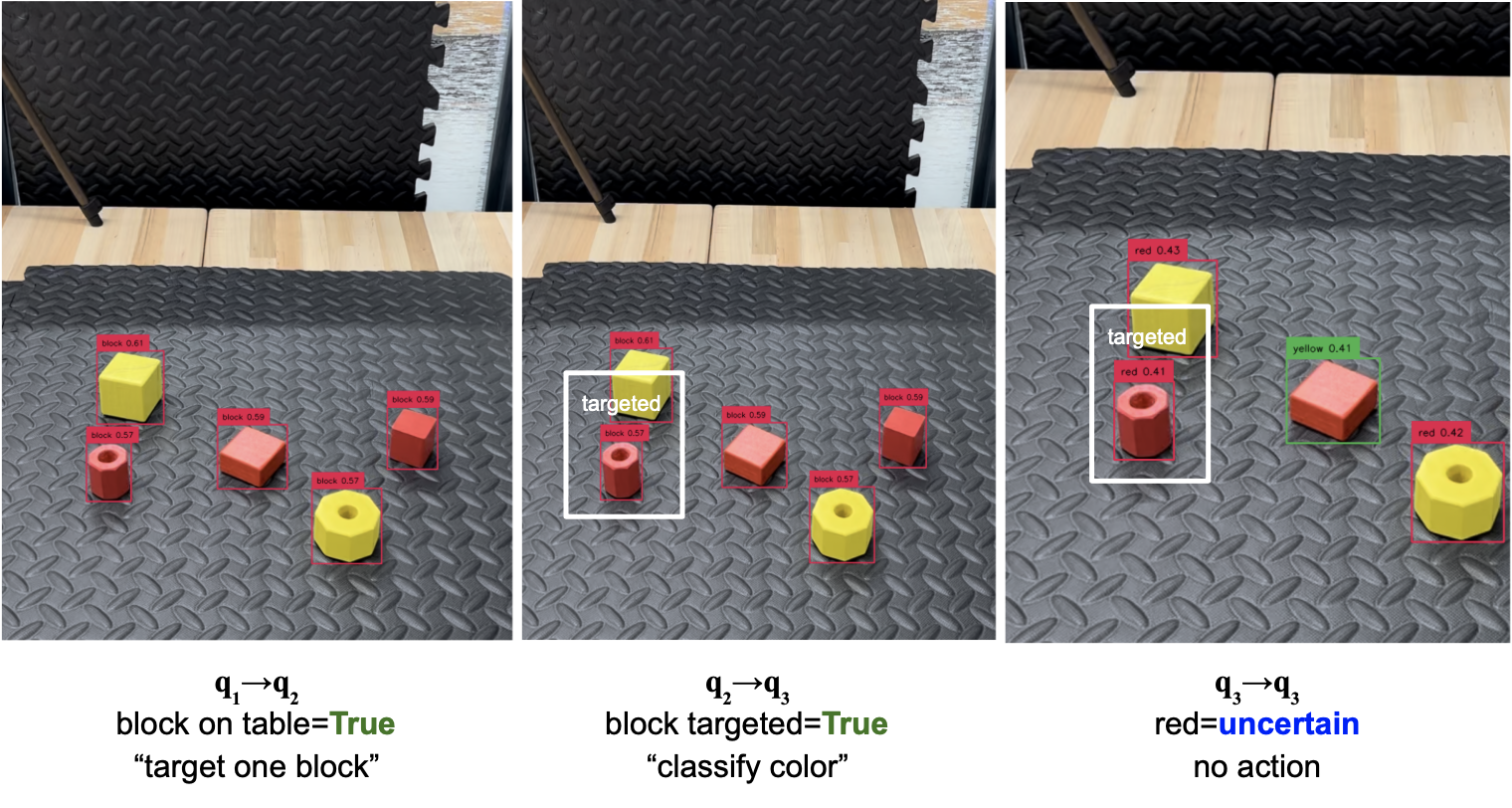}
    \includegraphics[height=0.3\linewidth]{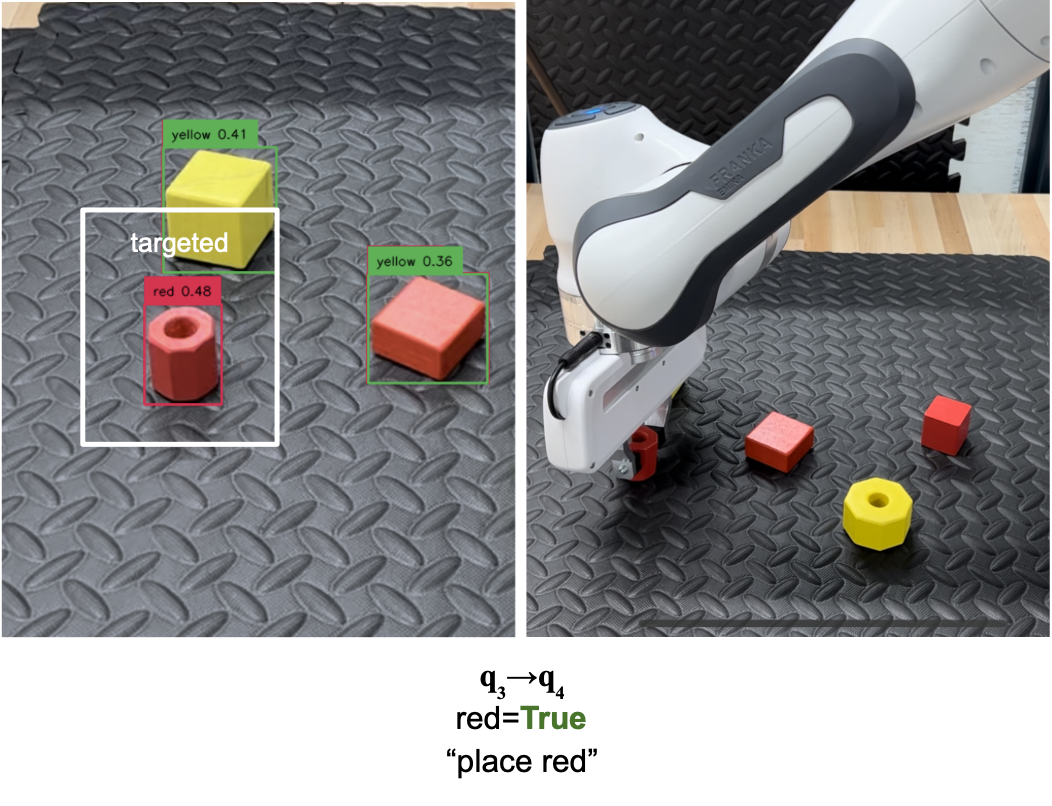}
    \caption{An example of using a robot arm to remove red blocks on the table. The figures show the object detection results from the \gls{vlm} in red and green boxes. We show the proposition evaluation results and list the state transitions and actions that are taken under the evaluated propositions.}
    \label{fig: arm-obs}
\end{figure*}

\section{Empirical Demonstration}
We illustrate the \GLMtoFSA{} algorithm and the grounding method \grounding{} with proof-of-concept examples on different domains. We also collect image observations from real-world environments to demonstrate the controller's behaviors in realistic deployment settings. We use the current state-of-the-art multimodal pretrained models \GPT{} \citep{openai2023gpt4} and \gls{vlm} \citep{kirillov2023segmentanything, liu2023grounding} to produce all the results in this section.

\subsection{Crossing Road Example}
We start the demonstration on a daily life task: cross the road. In this example, we construct a controller that can handle the crossing-road task at the traffic light. 

\paragraph{Controller Construction}
First, we query \GPT{} for the steps of crossing the road at the crosswalk at a traffic light and obtain a list of steps in textual form:
\begin{lstlisting}[language=completion]
    <prompt>Steps for "cross the road at a traffic light"</prompt>
    <completion> 1. Approach the pedestrian crossing. </completion>
    <completion> 2. Wait for the traffic light to turn green. </completion> 
    <completion> 3. Cross the road. </completion>
\end{lstlisting}
Next, we query GPT-4 again to define the verb phrases in PDDL and extract the preconditions and effects:
\begin{lstlisting}[language=completion]
    <prompt> Define an action "approach pedestrian crossing" in PDDL </prompt>
    <completion> Action: Approach_pedestrian_crossing </completion>
    <completion> Precondition: (not (at_pedestrian_crossing)) </completion>
    <completion> Effect: (at_pedestrian_crossing)</completion>
    
    <prompt> Define an action "Traffic light turn green" in PDDL</prompt>
    <completion> Action: Traffic_light_turn_green </completion>
    <completion> Precondition: (not (traffic_light_is_green)) </completion>
    <completion> Effect: (traffic_light_is_green)</completion>
    
    <prompt> Define an action "cross road" in PDDL</prompt>
    <completion> Action: Cross_road </completion>
    <completion> Precondition: (traffic_light_is_green) (at_pedestrian_crossing) </completion>
    <completion> Effect: (at other_side) </completion>
\end{lstlisting}

After we have the verb phrases with preconditions and effects in textual form, we follow the grammar in Table \ref{tab: grammar} to transform each step into a state and its outgoing transitions. We get an FSA that represents the controller by connecting all the states with the transitions, as presented in Figure \ref{fig: cross}.

\paragraph{Grounding and Verification}
We use the \gls{vlm} to evaluate the input symbols in the real-world task environment. The \gls{vlm} takes an image and a set of propositions in textual form as inputs and classifies which propositions match the image. 
A proposition matches an image if the object or scenario described by the proposition appears in the image.

In the grounding procedure, we apply Algorithm \ref{alg: eval-prop} with a true threshold $t=0.45$ and a false threshold $f=0.2$. A proposition will be evaluated as \UNC{} if the score is between 0.2 and 0.45. Under these thresholds, we have $\overline{p}_t = 0.983$, $\overline{p}_f = 0.975$, and $\mathbb{P}[\mathcal{A}_1=true]=0.975$ according to Figure \ref{fig: stats} and Theorem \ref{thm: prob}.
We also adjust the controller to adapt to the real-world environment with perceptual uncertainties, as presented in Figure \ref{fig: cross}.

Figure \ref{fig: cross-uncertain} shows an example of grounding the controller to the real-world environment with perceptual uncertainties. 
We highlight the second image from the left in the observation sequence in Figure \ref{fig: cross-uncertain}. Due to a confidence score of 0.4, the proposition ``traffic light is green" is evaluated to \UNC{}, which triggers a self-transition at state $q_{2}$, and no action is taken. Note that the \gls{vlm} misclassified the red light to the green light. If we do not consider perceptual uncertainties, the cross-road action may be triggered at the red light.

We use the model in Figure \ref{fig: cross-tl-model} to verify the controller with uncertainties against the specification $\Phi = \neg (cross road \land \neg green)$ to ensure safety. The controller satisfies the specification if Assumption 1 holds: $\mathcal{M} \otimes \Aut[controller] \models (\mathcal{A}_1 \implies \Phi)$. Then, according to Theorem \ref{thm: prob2}, the probability of the controller "never performing the cross-road action when the traffic light is not green" is at least $0.975^N$. Note that $N$ is the number of proposition evaluations that are certain during the grounding procedure ($N = 5$ in the example in Figure \ref{fig: cross-uncertain}. Hence, we have provided a guarantee for the safety of the controller. 

\subsection{Robot Arm Manipulation}
We follow \GLMtoFSA{} to construct a controller for the task ``use a robot arm to remove all the red blocks off the table." We show how we use the \gls{vlm} to perceive the operating environment and make decisions accordingly.

\begin{figure}[t]
    \centering
    \begin{tikzpicture}[thick,scale=.7, node distance=2.2cm, every node/.style={transform shape}]
	\node[state,initial] (1) at (0, 4) {\Large $q_{1}$};
	\node[state] (2) at (0, 0) {\Large $q_{2}$};
    \node[state] (3) at (4, 0) {\Large $q_{3}$};
	\node[state] (4) at (9, 0) {\Large $q_{4}$};

	\draw[->, shorten >=1pt] (1) to[loop above] node[align=center] {\small (\UNC{} $\vee \neg$ \textit{block on table}, $\noop$)} ();
    \draw[->, shorten >=1pt, sloped] (1) to[left] node[above, align=center] {\small (\textit{block on table}, \\ ``target one block")} (2);

    \draw[->, shorten >=1pt] (2) to[loop below] node[align=center] {\small ( \UNC{} $\vee \neg$ \textit{targeted}, $\noop$)} ();
    \draw[->, shorten >=1pt, sloped] (2) to[left] node[above, align=center] {\small (\textit{targeted}, \\ ``classify color")} (3);

    \draw[->, shorten >=1pt] (3) to[loop below] node[align=center] {\small ( \textit{\UNC{} $\vee \neg$ red $\land \neg$ yellow}, $\noop$)} ();
    \draw[->, shorten >=1pt, sloped] (3) to[bend left] node[above, align=center] {\small (\textit{red}, \\ ``place block")} (4);
    \draw[->, shorten >=1pt, sloped] (3) to[bend right] node[above, align=center] {\small (\textit{yellow}, \\ ``leave block")} (4);
    
    \draw[->, shorten >=1pt] (4) to[bend right] node[above, align=center] {\small (True, $\noop$)} (1);
\end{tikzpicture}
    \caption{The controller for a task "using a robot arm to remove all the red blocks on the table" with the consideration of perceptual uncertainties.}
    \label{fig: arm}
\end{figure}
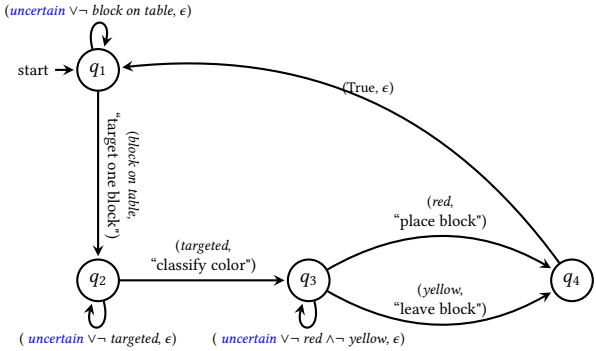

\paragraph{Controller Construction}
In this example, we assume the user has some prior knowledge of the task, such as some basic knowledge of the environment and the admissible actions of the robot arm. The user thus queries \GPT{} with the following prompt:
\begin{lstlisting}[language=completion]
    <prompt> Task: place all the red blocks off the table.</prompt>
    <prompt>Environment: there are unknown numbers of red blocks and yellow blocks on the table initially. Someone may randomly add a red block or yellow block to the table.</prompt>
    <prompt>Steps for achieving the task:</prompt>
    <completion> 1. Target one block on the table. </completion>
    <completion> 2. Classify the color of the targeted block. </completion>
    <completion> 3. If the block is red, place it from the table to an off-table location (B). If the block is yellow, leave it on the table.</completion>
    <completion> 4. Go to step 1. </completion>

    <prompt> 1. Define an action "target one block" in PDDL.</prompt>
    <completion> Action: target-one-block </completion>
    <completion> Parameters: () </completion>
    <completion> Precondition: (block_on_table) </completion>
    <completion> Effect: (and (block_targeted))</completion>
    
    <prompt> 2. Define an action "classify the color of the targeted block" in PDDL.</prompt>
    <completion> Action: classify-color </completion>
    <completion> Parameters: () </completion>
    <completion> Precondition: (block_targeted) </completion>
    <completion> Effect: (or (block_red) (block_yellow))</completion>
    
    <prompt> 3. Define an action "place the block" in PDDL.</prompt>
    <completion> Action: place-block </completion>
    <completion> Parameters: () </completion>
    <completion> Precondition: (and (block_targeted) (block_red)) </completion>
    <completion> Effect: (not (block_targeted))</completion>
    
    <prompt> 3. Define an action "leave the block" in PDDL.</prompt>
    <completion> Action: leave-block </completion>
    <completion> Parameters: (?b - block ?t - table) </completion>
    <completion> Precondition: (and (block_targeted) (block_yellow)) </completion>
    <completion> Effect: (not (block_targeted))</completion>
\end{lstlisting}
We present the constructed controller in Figure \ref{fig: arm}.

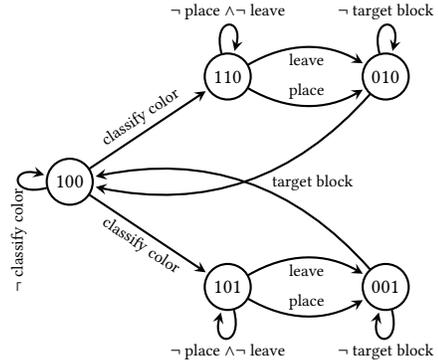
\begin{figure}[!t]
    \centering
    \begin{tikzpicture}[
    scale=.7,
    node distance=2.2cm,
    thick,
    every node/.append style={transform shape},
]

\node[state] (q0)
    at (0,0)
    {\Large $100$};
\node[state] (q1)
    at (3,2)
    {\Large $110$};
\node[state] (q2)
    at (3,-2)
    {\Large $101$};
\node[state] (q3)
    at (6,2)
    {\Large $010$};
\node[state] (q4)
    at (6,-2)
    {\Large $001$};

\path[->,sloped]

(q0) 
edge[loop left] node[]
    {$\neg$ classify color}
    ()
edge[] node[above]
    { classify color }
    (q1)
edge[] node[below]
    { classify color }
    (q2)

(q1) 
edge[loop above] node[sloped=false]
    {$\neg$ place $\land \neg$ leave }
    ()
edge[bend right] node[above]
    { place }
    (q3)
edge[bend left] node[below]
    { leave }
    (q3)
    
(q2) 
edge[loop below] node[sloped=false]
    { $\neg$ place $\land \neg$ leave }
    ()
edge[bend right] node[above]
    { place }
    (q4)
edge[bend left] node[below]
    { leave }
    (q4)

(q3) 
edge[loop above] node[sloped=false]
    {$\neg$ target block}
    ()
edge[bend left] node[]
    {  }
    (q0)

(q4) 
edge[loop below] node[sloped=false]
    {$\neg$ target block}
    ()
edge[bend right] node[right, sloped=false]
    { $\quad$ target block }
    (q0)
;

\end{tikzpicture}
    \caption{Transition system that represents the environment of the robot arm task. The environment requires the agent to target a block before classifying its color and to either place or leave the block only after the color is classified. The label on the state indicates the value of propositions (``block targeted," ``red," ``yellow").}
    \label{fig: arm-model}
\end{figure}

\paragraph{Grounding and Verification}
Next, we verify the controller in Figure \ref{fig: arm}. Suppose a model $\mathcal{M}$ is provided from some independent knowledge source and presented in Figure \ref{fig: arm-model}.
We want to guarantee the robot arm never accidentally places a yellow block outside the table. Hence we define the temporal logic specification
\begin{center}
    $\Phi = \neg place \land yellow$.
\end{center}
Recall that the trajectory is defined over the union of the set of actions and the set of effects. We use this model to verify that the robot arm controller satisfies the specification $\Phi$, under Assumption 1. The model-checking result indicates that the controller, when implemented in the model, satisfies $\Phi$ given Assumption 1 holds.

We again use the \gls{vlm} as the perception model to ground the controller from Figure \ref{fig: arm} to the operating environment. We set the true threshold and false threshold in Algorithm \ref{alg: eval-prop} to 0.45 and 0.2, respectively. Therefore, the probability $\mathbb{P}[\mathcal{A}_1=true]$ is 0.975. 
Figure \ref{fig: arm-obs} shows a full iteration of the controller ($q_1 \rightarrow q_2 \rightarrow q_3 \rightarrow q_4$). In the example from Figure \ref{fig: arm-obs}, the probability of Assumption 1 always holding is $0.975^3$. Therefore, the probability of $\Phi$ being satisfied in this example is also $0.975^3$.

\section{Conclusion}
We provide a proof-of-concept for the automatic construction of an automaton-based task controller of task knowledge from \glspl{glm} and the grounding of the controller to physical task environments.
We propose an algorithm named \GLMtoFSA{} that fills the gap between the textual outputs of generative models and sequential decision-making in the aspects of synthesis, verification, grounding, and perception.
The algorithm synthesizes automaton-based controllers from the text-based descriptions of task-relevant knowledge that are obtained from a \gls{glm}. Such automaton-based controllers can be verified against user-provided specifications over models representing the task environments or task knowledge from other independent sources.
Additionally, we develop a grounding method \grounding{} that grounds the automaton-based controllers to physical environments. It uses vision-language models to interpret visual observations from the task environment and implements control logic based on the observations. \grounding{} utilizes the confidence scores returned by the vision-language models to ensure safety under perceptual uncertainties.
Experimental results demonstrate the capabilities of \GLMtoFSA{} and \grounding{} on synthesis, verification, grounding, and perception.

\paragraph{Future Directions}
We have developed the algorithm to create formal representations of textual task knowledge and to ground those abstract representations in the physical environment through visual perceptions. As one future direction, we can develop an active perception method to actively search for the desired objects rather than having a fixed-angle camera.


\begin{acks}
This research was supported in part by the Office of Naval Research (ONR) under Grant N00014-22-1-2254 and in part by the National Science Foundation (NSF) under Grants NSF 1652113 and NSF 2211432.
\end{acks}



\bibliographystyle{ACM-Reference-Format}
\balance
\bibliography{references}


\begin{thebibliography}{36}


\ifx \showCODEN    \undefined \def \showCODEN     #1{\unskip}     \fi
\ifx \showDOI      \undefined \def \showDOI       #1{#1}\fi
\ifx \showISBNx    \undefined \def \showISBNx     #1{\unskip}     \fi
\ifx \showISBNxiii \undefined \def \showISBNxiii  #1{\unskip}     \fi
\ifx \showISSN     \undefined \def \showISSN      #1{\unskip}     \fi
\ifx \showLCCN     \undefined \def \showLCCN      #1{\unskip}     \fi
\ifx \shownote     \undefined \def \shownote      #1{#1}          \fi
\ifx \showarticletitle \undefined \def \showarticletitle #1{#1}   \fi
\ifx \showURL      \undefined \def \showURL       {\relax}        \fi
\providecommand\bibfield[2]{#2}
\providecommand\bibinfo[2]{#2}
\providecommand\natexlab[1]{#1}
\providecommand\showeprint[2][]{arXiv:#2}

\bibitem[\protect\citeauthoryear{Biggar and Zamani}{Biggar and Zamani}{2020}]%
        {Biggar2020AFF}
\bibfield{author}{\bibinfo{person}{Oliver Biggar} {and}
  \bibinfo{person}{Mohammad Zamani}.} \bibinfo{year}{2020}\natexlab{}.
\newblock \showarticletitle{A Framework for Formal Verification of Behavior
  Trees with Linear Temporal Logic}.
\newblock \bibinfo{journal}{\emph{{IEEE} Robotics and Automation Letters}}
  \bibinfo{volume}{5}, \bibinfo{number}{2} (\bibinfo{year}{2020}),
  \bibinfo{pages}{2341--2348}.
\newblock
\urldef\tempurl%
\url{https://doi.org/10.1109/LRA.2020.2970634}
\showDOI{\tempurl}


\bibitem[\protect\citeauthoryear{Brown, Mann, Ryder, Subbiah, Kaplan, Dhariwal,
  Neelakantan, Shyam, Sastry, Askell, Agarwal, Herbert{-}Voss, Krueger,
  Henighan, Child, Ramesh, Ziegler, Wu, Winter, Hesse, Chen, Sigler, Litwin,
  Gray, Chess, Clark, Berner, McCandlish, Radford, Sutskever, and Amodei}{Brown
  et~al\mbox{.}}{2020}]%
        {brown2020GPT3}
\bibfield{author}{\bibinfo{person}{Tom~B. Brown}, \bibinfo{person}{Benjamin
  Mann}, \bibinfo{person}{Nick Ryder}, \bibinfo{person}{Melanie Subbiah},
  \bibinfo{person}{Jared Kaplan}, \bibinfo{person}{Prafulla Dhariwal},
  \bibinfo{person}{Arvind Neelakantan}, \bibinfo{person}{Pranav Shyam},
  \bibinfo{person}{Girish Sastry}, \bibinfo{person}{Amanda Askell},
  \bibinfo{person}{Sandhini Agarwal}, \bibinfo{person}{Ariel Herbert{-}Voss},
  \bibinfo{person}{Gretchen Krueger}, \bibinfo{person}{Tom Henighan},
  \bibinfo{person}{Rewon Child}, \bibinfo{person}{Aditya Ramesh},
  \bibinfo{person}{Daniel~M. Ziegler}, \bibinfo{person}{Jeffrey Wu},
  \bibinfo{person}{Clemens Winter}, \bibinfo{person}{Christopher Hesse},
  \bibinfo{person}{Mark Chen}, \bibinfo{person}{Eric Sigler},
  \bibinfo{person}{Mateusz Litwin}, \bibinfo{person}{Scott Gray},
  \bibinfo{person}{Benjamin Chess}, \bibinfo{person}{Jack Clark},
  \bibinfo{person}{Christopher Berner}, \bibinfo{person}{Sam McCandlish},
  \bibinfo{person}{Alec Radford}, \bibinfo{person}{Ilya Sutskever}, {and}
  \bibinfo{person}{Dario Amodei}.} \bibinfo{year}{2020}\natexlab{}.
\newblock \showarticletitle{Language Models are Few-shot Learners}.
\newblock \bibinfo{journal}{\emph{Advances in Neural Information Processing
  Systems}}  \bibinfo{volume}{33} (\bibinfo{year}{2020}),
  \bibinfo{pages}{1877--1901}.
\newblock


\bibitem[\protect\citeauthoryear{Chang, Lambert, Sangkloy, Singh, Bak,
  Hartnett, Wang, Carr, Lucey, Ramanan, and Hays}{Chang et~al\mbox{.}}{2019}]%
        {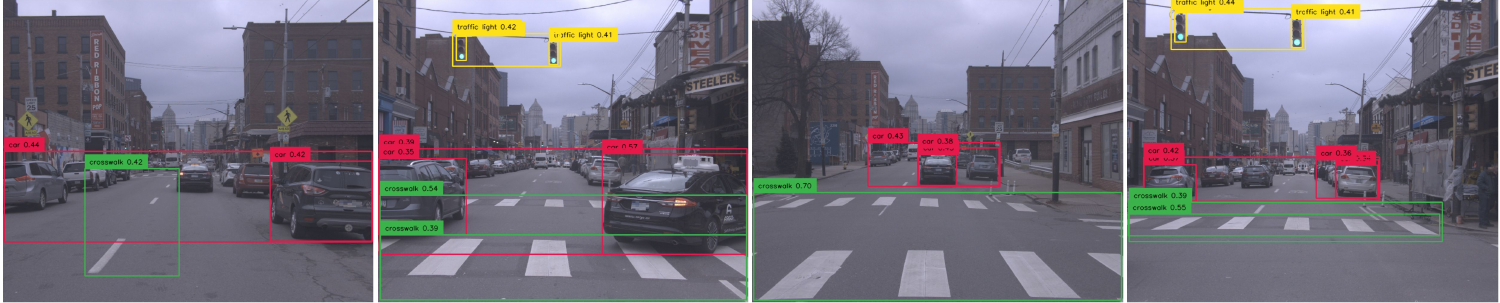}
\bibfield{author}{\bibinfo{person}{Ming{-}Fang Chang}, \bibinfo{person}{John
  Lambert}, \bibinfo{person}{Patsorn Sangkloy}, \bibinfo{person}{Jagjeet
  Singh}, \bibinfo{person}{Slawomir Bak}, \bibinfo{person}{Andrew Hartnett},
  \bibinfo{person}{De Wang}, \bibinfo{person}{Peter Carr},
  \bibinfo{person}{Simon Lucey}, \bibinfo{person}{Deva Ramanan}, {and}
  \bibinfo{person}{James Hays}.} \bibinfo{year}{2019}\natexlab{}.
\newblock \showarticletitle{Argoverse: 3D Tracking and Forecasting With Rich
  Maps}. In \bibinfo{booktitle}{\emph{{IEEE} Conference on Computer Vision and
  Pattern Recognition}}. \bibinfo{publisher}{Computer Vision Foundation /
  {IEEE}}, \bibinfo{address}{Long Beach, CA, USA}, \bibinfo{pages}{8748--8757}.
\newblock


\bibitem[\protect\citeauthoryear{Chen, Tworek, Jun, Yuan, Ponde, Kaplan,
  Edwards, Burda, Joseph, Brockman, Ray, Puri, and et~al.}{Chen
  et~al\mbox{.}}{2021}]%
        {codex}
\bibfield{author}{\bibinfo{person}{Mark Chen}, \bibinfo{person}{Jerry Tworek},
  \bibinfo{person}{Heewoo Jun}, \bibinfo{person}{Qiming Yuan},
  \bibinfo{person}{Henrique Ponde}, \bibinfo{person}{Jared Kaplan},
  \bibinfo{person}{Harrison Edwards}, \bibinfo{person}{Yura Burda},
  \bibinfo{person}{Nicholas Joseph}, \bibinfo{person}{Greg Brockman},
  \bibinfo{person}{Alex Ray}, \bibinfo{person}{Raul Puri}, {and}
  \bibinfo{person}{et al.}} \bibinfo{year}{2021}\natexlab{}.
\newblock \bibinfo{title}{Evaluating Large Language Models Trained on Code}.
\newblock
\newblock


\bibitem[\protect\citeauthoryear{Chowdhery, Narang, Devlin, Bosma, Mishra, and
  et~al.}{Chowdhery et~al\mbox{.}}{2022}]%
        {Chowdhery2022PaLMSL}
\bibfield{author}{\bibinfo{person}{Aakanksha Chowdhery},
  \bibinfo{person}{Sharan Narang}, \bibinfo{person}{Jacob Devlin},
  \bibinfo{person}{Maarten Bosma}, \bibinfo{person}{Gaurav Mishra}, {and}
  \bibinfo{person}{et al.}} \bibinfo{year}{2022}\natexlab{}.
\newblock \bibinfo{title}{PaLM: Scaling Language Modeling with Pathways}.
\newblock
\newblock


\bibitem[\protect\citeauthoryear{Cimatti, Clarke, Giunchiglia, Giunchiglia,
  Pistore, Roveri, Sebastiani, and Tacchella}{Cimatti et~al\mbox{.}}{2002}]%
        {Cimatti2002NuSMV}
\bibfield{author}{\bibinfo{person}{Alessandro Cimatti},
  \bibinfo{person}{Edmund~M. Clarke}, \bibinfo{person}{Enrico Giunchiglia},
  \bibinfo{person}{Fausto Giunchiglia}, \bibinfo{person}{Marco Pistore},
  \bibinfo{person}{Marco Roveri}, \bibinfo{person}{Roberto Sebastiani}, {and}
  \bibinfo{person}{Armando Tacchella}.} \bibinfo{year}{2002}\natexlab{}.
\newblock \showarticletitle{Nu{SMV} 2: An OpenSource Tool for Symbolic Model
  Checking}. In \bibinfo{booktitle}{\emph{Computer Aided Verification}}
  \emph{(\bibinfo{series}{Lecture Notes in Computer Science},
  Vol.~\bibinfo{volume}{2404})}. \bibinfo{publisher}{Springer},
  \bibinfo{pages}{359--364}.
\newblock
\urldef\tempurl%
\url{https://doi.org/10.1007/3-540-45657-0\_29}
\showDOI{\tempurl}


\bibitem[\protect\citeauthoryear{Devlin, Chang, Lee, and Toutanova}{Devlin
  et~al\mbox{.}}{2019}]%
        {bert}
\bibfield{author}{\bibinfo{person}{Jacob Devlin}, \bibinfo{person}{Ming{-}Wei
  Chang}, \bibinfo{person}{Kenton Lee}, {and} \bibinfo{person}{Kristina
  Toutanova}.} \bibinfo{year}{2019}\natexlab{}.
\newblock \showarticletitle{{BERT:} Pre-training of Deep Bidirectional
  Transformers for Language Understanding}. In
  \bibinfo{booktitle}{\emph{Conference of the North American Chapter of the
  Association for Computational Linguistics: Human Language Technologies}},
  \bibfield{editor}{\bibinfo{person}{Jill Burstein}, \bibinfo{person}{Christy
  Doran}, {and} \bibinfo{person}{Thamar Solorio}} (Eds.).
  \bibinfo{publisher}{Association for Computational Linguistics},
  \bibinfo{address}{Minneapolis, MN, USA}, \bibinfo{pages}{4171--4186}.
\newblock


\bibitem[\protect\citeauthoryear{Gu, Lin, Kuo, and Cui}{Gu
  et~al\mbox{.}}{2022}]%
        {GuLKC22}
\bibfield{author}{\bibinfo{person}{Xiuye Gu}, \bibinfo{person}{Tsung{-}Yi Lin},
  \bibinfo{person}{Weicheng Kuo}, {and} \bibinfo{person}{Yin Cui}.}
  \bibinfo{year}{2022}\natexlab{}.
\newblock \showarticletitle{Open-vocabulary Object Detection via Vision and
  Language Knowledge Distillation}. In \bibinfo{booktitle}{\emph{International
  Conference on Learning Representations}}.
  \bibinfo{publisher}{OpenReview.net}, \bibinfo{address}{Virtual},
  \bibinfo{pages}{1--20}.
\newblock


\bibitem[\protect\citeauthoryear{Haslum, Lipovetzky, Magazzeni, and
  Muise}{Haslum et~al\mbox{.}}{2019}]%
        {pddl}
\bibfield{author}{\bibinfo{person}{Patrik Haslum}, \bibinfo{person}{Nir
  Lipovetzky}, \bibinfo{person}{Daniele Magazzeni}, {and}
  \bibinfo{person}{Christian Muise}.} \bibinfo{year}{2019}\natexlab{}.
\newblock \bibinfo{booktitle}{\emph{An Introduction to the Planning Domain
  Definition Language}}.
\newblock \bibinfo{publisher}{Morgan {\&} Claypool Publishers}.
\newblock


\bibitem[\protect\citeauthoryear{He, Fang, Wang, and Song}{He
  et~al\mbox{.}}{2022}]%
        {He2022AcquiringAM}
\bibfield{author}{\bibinfo{person}{Mutian He}, \bibinfo{person}{Tianqing Fang},
  \bibinfo{person}{Weiqi Wang}, {and} \bibinfo{person}{Yangqiu Song}.}
  \bibinfo{year}{2022}\natexlab{}.
\newblock \bibinfo{title}{Acquiring and Modelling Abstract Commonsense
  Knowledge via Conceptualization}.
\newblock
\newblock


\bibitem[\protect\citeauthoryear{Huang, Abbeel, Pathak, and Mordatch}{Huang
  et~al\mbox{.}}{2022a}]%
        {huang2022language}
\bibfield{author}{\bibinfo{person}{Wenlong Huang}, \bibinfo{person}{Pieter
  Abbeel}, \bibinfo{person}{Deepak Pathak}, {and} \bibinfo{person}{Igor
  Mordatch}.} \bibinfo{year}{2022}\natexlab{a}.
\newblock \showarticletitle{Language Models as Zero-Shot Planners: Extracting
  Actionable Knowledge for Embodied Agents}. In
  \bibinfo{booktitle}{\emph{International Conference on Machine Learning}}
  \emph{(\bibinfo{series}{Proceedings of Machine Learning Research},
  Vol.~\bibinfo{volume}{162})}. \bibinfo{publisher}{{PMLR}},
  \bibinfo{address}{Baltimore, Maryland, {USA}}, \bibinfo{pages}{9118--9147}.
\newblock


\bibitem[\protect\citeauthoryear{Huang, Xia, Xiao, Chan, Liang, Florence, Zeng,
  Tompson, Mordatch, Chebotar, Sermanet, Jackson, Brown, Luu, Levine, Hausman,
  and Ichter}{Huang et~al\mbox{.}}{2022b}]%
        {huang2022inner}
\bibfield{author}{\bibinfo{person}{Wenlong Huang}, \bibinfo{person}{Fei Xia},
  \bibinfo{person}{Ted Xiao}, \bibinfo{person}{Harris Chan},
  \bibinfo{person}{Jacky Liang}, \bibinfo{person}{Pete Florence},
  \bibinfo{person}{Andy Zeng}, \bibinfo{person}{Jonathan Tompson},
  \bibinfo{person}{Igor Mordatch}, \bibinfo{person}{Yevgen Chebotar},
  \bibinfo{person}{Pierre Sermanet}, \bibinfo{person}{Tomas Jackson},
  \bibinfo{person}{Noah Brown}, \bibinfo{person}{Linda Luu},
  \bibinfo{person}{Sergey Levine}, \bibinfo{person}{Karol Hausman}, {and}
  \bibinfo{person}{Brian Ichter}.} \bibinfo{year}{2022}\natexlab{b}.
\newblock \showarticletitle{Inner Monologue: Embodied Reasoning through
  Planning with Language Models}. In \bibinfo{booktitle}{\emph{Conference on
  Robot Learning}} \emph{(\bibinfo{series}{Proceedings of Machine Learning
  Research}, Vol.~\bibinfo{volume}{205})}. \bibinfo{publisher}{PLMR},
  \bibinfo{address}{Auckland, New Zealand}, \bibinfo{pages}{1769--1782}.
\newblock


\bibitem[\protect\citeauthoryear{Ichter, Brohan, Chebotar, Finn, Hausman,
  Herzog, Ho, Ibarz, Irpan, Jang, Julian, Kalashnikov, Levine, Lu, Parada, Rao,
  Sermanet, Toshev, Vanhoucke, Xia, Xiao, Xu, Yan, Brown, Ahn, Cortes, Sievers,
  Tan, Xu, Reyes, Rettinghouse, Quiambao, Pastor, Luu, Lee, Kuang, Jesmonth,
  Joshi, Jeffrey, Ruano, Hsu, Gopalakrishnan, David, Zeng, and Fu}{Ichter
  et~al\mbox{.}}{2022}]%
        {brohan2023can}
\bibfield{author}{\bibinfo{person}{Brian Ichter}, \bibinfo{person}{Anthony
  Brohan}, \bibinfo{person}{Yevgen Chebotar}, \bibinfo{person}{Chelsea Finn},
  \bibinfo{person}{Karol Hausman}, \bibinfo{person}{Alexander Herzog},
  \bibinfo{person}{Daniel Ho}, \bibinfo{person}{Julian Ibarz},
  \bibinfo{person}{Alex Irpan}, \bibinfo{person}{Eric Jang},
  \bibinfo{person}{Ryan Julian}, \bibinfo{person}{Dmitry Kalashnikov},
  \bibinfo{person}{Sergey Levine}, \bibinfo{person}{Yao Lu},
  \bibinfo{person}{Carolina Parada}, \bibinfo{person}{Kanishka Rao},
  \bibinfo{person}{Pierre Sermanet}, \bibinfo{person}{Alexander Toshev},
  \bibinfo{person}{Vincent Vanhoucke}, \bibinfo{person}{Fei Xia},
  \bibinfo{person}{Ted Xiao}, \bibinfo{person}{Peng Xu},
  \bibinfo{person}{Mengyuan Yan}, \bibinfo{person}{Noah Brown},
  \bibinfo{person}{Michael Ahn}, \bibinfo{person}{Omar Cortes},
  \bibinfo{person}{Nicolas Sievers}, \bibinfo{person}{Clayton Tan},
  \bibinfo{person}{Sichun Xu}, \bibinfo{person}{Diego Reyes},
  \bibinfo{person}{Jarek Rettinghouse}, \bibinfo{person}{Jornell Quiambao},
  \bibinfo{person}{Peter Pastor}, \bibinfo{person}{Linda Luu},
  \bibinfo{person}{Kuang{-}Huei Lee}, \bibinfo{person}{Yuheng Kuang},
  \bibinfo{person}{Sally Jesmonth}, \bibinfo{person}{Nikhil~J. Joshi},
  \bibinfo{person}{Kyle Jeffrey}, \bibinfo{person}{Rosario~Jauregui Ruano},
  \bibinfo{person}{Jasmine Hsu}, \bibinfo{person}{Keerthana Gopalakrishnan},
  \bibinfo{person}{Byron David}, \bibinfo{person}{Andy Zeng}, {and}
  \bibinfo{person}{Chuyuan~Kelly Fu}.} \bibinfo{year}{2022}\natexlab{}.
\newblock \showarticletitle{Do As {I} Can, Not As {I} Say: Grounding Language
  in Robotic Affordances}. In \bibinfo{booktitle}{\emph{Conference on Robot
  Learning}} \emph{(\bibinfo{series}{Proceedings of Machine Learning Research},
  Vol.~\bibinfo{volume}{205})}. \bibinfo{publisher}{PLMR},
  \bibinfo{address}{Auckland, New Zealand}, \bibinfo{pages}{287--318}.
\newblock


\bibitem[\protect\citeauthoryear{Kirillov, Mintun, Ravi, Mao, Rolland,
  Gustafson, Xiao, Whitehead, Berg, Lo, Doll{\'a}r, and Girshick}{Kirillov
  et~al\mbox{.}}{2023}]%
        {kirillov2023segmentanything}
\bibfield{author}{\bibinfo{person}{Alexander Kirillov}, \bibinfo{person}{Eric
  Mintun}, \bibinfo{person}{Nikhila Ravi}, \bibinfo{person}{Hanzi Mao},
  \bibinfo{person}{Chloe Rolland}, \bibinfo{person}{Laura Gustafson},
  \bibinfo{person}{Tete Xiao}, \bibinfo{person}{Spencer Whitehead},
  \bibinfo{person}{Alexander~C. Berg}, \bibinfo{person}{Wan-Yen Lo},
  \bibinfo{person}{Piotr Doll{\'a}r}, {and} \bibinfo{person}{Ross Girshick}.}
  \bibinfo{year}{2023}\natexlab{}.
\newblock \bibinfo{title}{Segment Anything}.
\newblock
\newblock


\bibitem[\protect\citeauthoryear{Li, Zhang, Zhang, Yang, Li, Zhong, Wang, Yuan,
  Zhang, Hwang, Chang, and Gao}{Li et~al\mbox{.}}{2022}]%
        {LiZZYLZWYZHCG22}
\bibfield{author}{\bibinfo{person}{Liunian~Harold Li},
  \bibinfo{person}{Pengchuan Zhang}, \bibinfo{person}{Haotian Zhang},
  \bibinfo{person}{Jianwei Yang}, \bibinfo{person}{Chunyuan Li},
  \bibinfo{person}{Yiwu Zhong}, \bibinfo{person}{Lijuan Wang},
  \bibinfo{person}{Lu Yuan}, \bibinfo{person}{Lei Zhang},
  \bibinfo{person}{Jenq{-}Neng Hwang}, \bibinfo{person}{Kai{-}Wei Chang}, {and}
  \bibinfo{person}{Jianfeng Gao}.} \bibinfo{year}{2022}\natexlab{}.
\newblock \showarticletitle{Grounded Language-Image Pre-training}. In
  \bibinfo{booktitle}{\emph{Conference on Computer Vision and Pattern
  Recognition}}. \bibinfo{publisher}{{IEEE}}, \bibinfo{address}{New Orleans,
  LA, USA}, \bibinfo{pages}{10955--10965}.
\newblock


\bibitem[\protect\citeauthoryear{Lin, Huang, Liu, Gu, Sommerer, and Ren}{Lin
  et~al\mbox{.}}{2023b}]%
        {LinHLGS023}
\bibfield{author}{\bibinfo{person}{Bill~Yuchen Lin}, \bibinfo{person}{Chengsong
  Huang}, \bibinfo{person}{Qian Liu}, \bibinfo{person}{Wenda Gu},
  \bibinfo{person}{Sam Sommerer}, {and} \bibinfo{person}{Xiang Ren}.}
  \bibinfo{year}{2023}\natexlab{b}.
\newblock \showarticletitle{On Grounded Planning for Embodied Tasks with
  Language Models}. In \bibinfo{booktitle}{\emph{{AAAI} Conference on
  Artificial Intelligence}}, \bibfield{editor}{\bibinfo{person}{Brian
  Williams}, \bibinfo{person}{Yiling Chen}, {and} \bibinfo{person}{Jennifer
  Neville}} (Eds.). \bibinfo{publisher}{{AAAI} Press},
  \bibinfo{address}{Washington, DC, USA}, \bibinfo{pages}{13192--13200}.
\newblock


\bibitem[\protect\citeauthoryear{Lin, Agia, Migimatsu, Pavone, and Bohg}{Lin
  et~al\mbox{.}}{2023a}]%
        {Lin2023Text2MotionFN}
\bibfield{author}{\bibinfo{person}{Kevin Lin}, \bibinfo{person}{Christopher
  Agia}, \bibinfo{person}{Toki Migimatsu}, \bibinfo{person}{Marco Pavone},
  {and} \bibinfo{person}{Jeannette Bohg}.} \bibinfo{year}{2023}\natexlab{a}.
\newblock \bibinfo{title}{Text2Motion: From Natural Language Instructions to
  Feasible Plans}.
\newblock
\newblock


\bibitem[\protect\citeauthoryear{Liu, Jiang, Zhang, Liu, Zhang, Biswas, and
  Stone}{Liu et~al\mbox{.}}{2023a}]%
        {Liu2023LLMPEL}
\bibfield{author}{\bibinfo{person}{B. Liu}, \bibinfo{person}{Yuqian Jiang},
  \bibinfo{person}{Xiaohan Zhang}, \bibinfo{person}{Qian Liu},
  \bibinfo{person}{Shiqi Zhang}, \bibinfo{person}{Joydeep Biswas}, {and}
  \bibinfo{person}{Peter Stone}.} \bibinfo{year}{2023}\natexlab{a}.
\newblock \bibinfo{title}{LLM+P: Empowering Large Language Models with Optimal
  Planning Proficiency}.
\newblock
\newblock


\bibitem[\protect\citeauthoryear{Liu, Zeng, Ren, Li, Zhang, Yang, Li, Yang, Su,
  Zhu, et~al\mbox{.}}{Liu et~al\mbox{.}}{2023b}]%
        {liu2023grounding}
\bibfield{author}{\bibinfo{person}{Shilong Liu}, \bibinfo{person}{Zhaoyang
  Zeng}, \bibinfo{person}{Tianhe Ren}, \bibinfo{person}{Feng Li},
  \bibinfo{person}{Hao Zhang}, \bibinfo{person}{Jie Yang},
  \bibinfo{person}{Chunyuan Li}, \bibinfo{person}{Jianwei Yang},
  \bibinfo{person}{Hang Su}, \bibinfo{person}{Jun Zhu}, {et~al\mbox{.}}}
  \bibinfo{year}{2023}\natexlab{b}.
\newblock \bibinfo{title}{Grounding {DINO:} Marrying {DINO} with Grounded
  Pre-Training for Open-Set Object Detection}.
\newblock
\newblock


\bibitem[\protect\citeauthoryear{Lu, Feng, Zhu, Xu, Wang, Eckstein, and
  Wang}{Lu et~al\mbox{.}}{2022}]%
        {Lu2022NeuroSymbolicCL}
\bibfield{author}{\bibinfo{person}{Yujie Lu}, \bibinfo{person}{Weixi Feng},
  \bibinfo{person}{Wanrong Zhu}, \bibinfo{person}{Wenda Xu},
  \bibinfo{person}{Xin~Eric Wang}, \bibinfo{person}{Miguel Eckstein}, {and}
  \bibinfo{person}{William~Yang Wang}.} \bibinfo{year}{2022}\natexlab{}.
\newblock \bibinfo{title}{Neuro-Symbolic Procedural Planning with Commonsense
  Prompting}.
\newblock
\newblock


\bibitem[\protect\citeauthoryear{Lu, Feng, Zhu, Xu, Wang, Eckstein, and
  Wang}{Lu et~al\mbox{.}}{2023a}]%
        {neural-symbolic}
\bibfield{author}{\bibinfo{person}{Yujie Lu}, \bibinfo{person}{Weixi Feng},
  \bibinfo{person}{Wanrong Zhu}, \bibinfo{person}{Wenda Xu},
  \bibinfo{person}{Xin~Eric Wang}, \bibinfo{person}{Miguel Eckstein}, {and}
  \bibinfo{person}{William~Yang Wang}.} \bibinfo{year}{2023}\natexlab{a}.
\newblock \showarticletitle{Neuro-Symbolic Procedural Planning with Commonsense
  Prompting}. In \bibinfo{booktitle}{\emph{International Conference on Learning
  Representations}}. \bibinfo{publisher}{OpenReview.net},
  \bibinfo{address}{Kigali, Rwanda}, \bibinfo{pages}{1--34}.
\newblock


\bibitem[\protect\citeauthoryear{Lu, Lu, Chen, Zhu, Wang, and Wang}{Lu
  et~al\mbox{.}}{2023b}]%
        {Lu2023MultimodalPP}
\bibfield{author}{\bibinfo{person}{Yujie Lu}, \bibinfo{person}{Pan Lu},
  \bibinfo{person}{Zhiyu Chen}, \bibinfo{person}{Wanrong Zhu},
  \bibinfo{person}{Xin~Eric Wang}, {and} \bibinfo{person}{William~Yang Wang}.}
  \bibinfo{year}{2023}\natexlab{b}.
\newblock \bibinfo{title}{Multimodal Procedural Planning via Dual Text-Image
  Prompting}.
\newblock
\newblock


\bibitem[\protect\citeauthoryear{OpenAI}{OpenAI}{2023}]%
        {openai2023gpt4}
\bibfield{author}{\bibinfo{person}{OpenAI}.} \bibinfo{year}{2023}\natexlab{}.
\newblock \bibinfo{title}{GPT-4 Technical Report}.
\newblock
\newblock
\showeprint[arxiv]{2303.08774}~[cs.CL]


\bibitem[\protect\citeauthoryear{Radford, Kim, Hallacy, Ramesh, Goh, Agarwal,
  Sastry, Askell, Mishkin, Clark, Krueger, and Sutskever}{Radford
  et~al\mbox{.}}{2021}]%
        {clip}
\bibfield{author}{\bibinfo{person}{Alec Radford}, \bibinfo{person}{Jong~Wook
  Kim}, \bibinfo{person}{Chris Hallacy}, \bibinfo{person}{Aditya Ramesh},
  \bibinfo{person}{Gabriel Goh}, \bibinfo{person}{Sandhini Agarwal},
  \bibinfo{person}{Girish Sastry}, \bibinfo{person}{Amanda Askell},
  \bibinfo{person}{Pamela Mishkin}, \bibinfo{person}{Jack Clark},
  \bibinfo{person}{Gretchen Krueger}, {and} \bibinfo{person}{Ilya Sutskever}.}
  \bibinfo{year}{2021}\natexlab{}.
\newblock \showarticletitle{Learning Transferable Visual Models From Natural
  Language Supervision}. In \bibinfo{booktitle}{\emph{International Conference
  on Machine Learning}} \emph{(\bibinfo{series}{Proceedings of Machine Learning
  Research}, Vol.~\bibinfo{volume}{139})},
  \bibfield{editor}{\bibinfo{person}{Marina Meila} {and} \bibinfo{person}{Tong
  Zhang}} (Eds.). \bibinfo{publisher}{{PMLR}}, \bibinfo{address}{Virtual},
  \bibinfo{pages}{8748--8763}.
\newblock


\bibitem[\protect\citeauthoryear{Redmon, Divvala, Girshick, and Farhadi}{Redmon
  et~al\mbox{.}}{2016}]%
        {yolo}
\bibfield{author}{\bibinfo{person}{Joseph Redmon},
  \bibinfo{person}{Santosh~Kumar Divvala}, \bibinfo{person}{Ross~B. Girshick},
  {and} \bibinfo{person}{Ali Farhadi}.} \bibinfo{year}{2016}\natexlab{}.
\newblock \showarticletitle{You Only Look Once: Unified, Real-Time Object
  Detection}. In \bibinfo{booktitle}{\emph{Conference on Computer Vision and
  Pattern Recognition}}. \bibinfo{publisher}{{IEEE} Computer Society},
  \bibinfo{address}{Las Vegas, NV, USA}, \bibinfo{pages}{779--788}.
\newblock


\bibitem[\protect\citeauthoryear{Ren, He, Girshick, and Sun}{Ren
  et~al\mbox{.}}{2017}]%
        {rcnn}
\bibfield{author}{\bibinfo{person}{Shaoqing Ren}, \bibinfo{person}{Kaiming He},
  \bibinfo{person}{Ross~B. Girshick}, {and} \bibinfo{person}{Jian Sun}.}
  \bibinfo{year}{2017}\natexlab{}.
\newblock \showarticletitle{Faster {R-CNN:} Towards Real-Time Object Detection
  with Region Proposal Networks}.
\newblock \bibinfo{journal}{\emph{{IEEE} Transaction Pattern Analysis Machine
  Intelligence}} \bibinfo{volume}{39}, \bibinfo{number}{6}
  (\bibinfo{year}{2017}), \bibinfo{pages}{1137--1149}.
\newblock


\bibitem[\protect\citeauthoryear{Rezaei and Reformat}{Rezaei and
  Reformat}{2022}]%
        {Rezaei2022UtilizingLM}
\bibfield{author}{\bibinfo{person}{Navid Rezaei} {and}
  \bibinfo{person}{Marek~Z. Reformat}.} \bibinfo{year}{2022}\natexlab{}.
\newblock \showarticletitle{Utilizing Language Models to Expand Vision-Based
  Commonsense Knowledge Graphs}.
\newblock \bibinfo{journal}{\emph{Symmetry}}  \bibinfo{volume}{14}
  (\bibinfo{year}{2022}), \bibinfo{pages}{1715}.
\newblock


\bibitem[\protect\citeauthoryear{Scao, Fan, Akiki, Pavlick, Ili'c, Hesslow,
  Castagn'e, Luccioni, Yvon, Gall{\'e}, and et~al.}{Scao et~al\mbox{.}}{2022}]%
        {Scao2022BLOOMA1}
\bibfield{author}{\bibinfo{person}{Teven~Le Scao}, \bibinfo{person}{Angela
  Fan}, \bibinfo{person}{Christopher Akiki}, \bibinfo{person}{Elizabeth-Jane
  Pavlick}, \bibinfo{person}{Suzana Ili'c}, \bibinfo{person}{Daniel Hesslow},
  \bibinfo{person}{Roman Castagn'e}, \bibinfo{person}{Alexandra~Sasha
  Luccioni}, \bibinfo{person}{Franccois Yvon}, \bibinfo{person}{Matthias
  Gall{\'e}}, {and} \bibinfo{person}{et al.}} \bibinfo{year}{2022}\natexlab{}.
\newblock \bibinfo{title}{BLOOM: A 176B-Parameter Open-Access Multilingual
  Language Model}.
\newblock
\newblock


\bibitem[\protect\citeauthoryear{Shah, Osinski, Ichter, and Levine}{Shah
  et~al\mbox{.}}{2022}]%
        {shah2022robotic}
\bibfield{author}{\bibinfo{person}{Dhruv Shah}, \bibinfo{person}{Blazej
  Osinski}, \bibinfo{person}{Brian Ichter}, {and} \bibinfo{person}{Sergey
  Levine}.} \bibinfo{year}{2022}\natexlab{}.
\newblock \showarticletitle{LM-Nav: Robotic Navigation with Large Pre-Trained
  Models of Language, Vision, and Action}. In
  \bibinfo{booktitle}{\emph{Conference on Robot Learning}}
  \emph{(\bibinfo{series}{Proceedings of Machine Learning Research},
  Vol.~\bibinfo{volume}{205})}. \bibinfo{publisher}{{PMLR}},
  \bibinfo{address}{Auckland, New Zealand}, \bibinfo{pages}{492--504}.
\newblock


\bibitem[\protect\citeauthoryear{Singh, Blukis, Mousavian, Goyal, Xu, Tremblay,
  Fox, Thomason, and Garg}{Singh et~al\mbox{.}}{2022}]%
        {Singh2022ProgPromptGS}
\bibfield{author}{\bibinfo{person}{Ishika Singh}, \bibinfo{person}{Valts
  Blukis}, \bibinfo{person}{Arsalan Mousavian}, \bibinfo{person}{Ankit Goyal},
  \bibinfo{person}{Danfei Xu}, \bibinfo{person}{Jonathan Tremblay},
  \bibinfo{person}{Dieter Fox}, \bibinfo{person}{Jesse Thomason}, {and}
  \bibinfo{person}{Animesh Garg}.} \bibinfo{year}{2022}\natexlab{}.
\newblock \bibinfo{title}{ProgPrompt: Generating Situated Robot Task Plans
  using Large Language Models}.
\newblock
\newblock


\bibitem[\protect\citeauthoryear{Smith, Patwary, Norick, LeGresley,
  Rajbhandari, Casper, Liu, Prabhumoye, Zerveas, Korthikanti, Zhang, Child,
  Aminabadi, Bernauer, Song, Shoeybi, He, Houston, Tiwary, and Catanzaro}{Smith
  et~al\mbox{.}}{2022}]%
        {Smith2022UsingDA}
\bibfield{author}{\bibinfo{person}{Shaden Smith}, \bibinfo{person}{Mostofa
  Patwary}, \bibinfo{person}{Brandon Norick}, \bibinfo{person}{Patrick
  LeGresley}, \bibinfo{person}{Samyam Rajbhandari}, \bibinfo{person}{Jared
  Casper}, \bibinfo{person}{Zhun Liu}, \bibinfo{person}{Shrimai Prabhumoye},
  \bibinfo{person}{George Zerveas}, \bibinfo{person}{Vijay~Anand Korthikanti},
  \bibinfo{person}{Elton Zhang}, \bibinfo{person}{Rewon Child},
  \bibinfo{person}{Reza~Yazdani Aminabadi}, \bibinfo{person}{Julie Bernauer},
  \bibinfo{person}{Xia Song}, \bibinfo{person}{Mohammad Shoeybi},
  \bibinfo{person}{Yuxiong He}, \bibinfo{person}{Michael Houston},
  \bibinfo{person}{Saurabh Tiwary}, {and} \bibinfo{person}{Bryan Catanzaro}.}
  \bibinfo{year}{2022}\natexlab{}.
\newblock \bibinfo{title}{Using DeepSpeed and Megatron to Train Megatron-Turing
  NLG 530B, A Large-Scale Generative Language Model}.
\newblock
\newblock


\bibitem[\protect\citeauthoryear{Song, Wu, Washington, Sadler, Chao, and
  Su}{Song et~al\mbox{.}}{2022}]%
        {Song2022LLMPlannerFG}
\bibfield{author}{\bibinfo{person}{Chan~Hee Song}, \bibinfo{person}{Jiaman Wu},
  \bibinfo{person}{Clay Washington}, \bibinfo{person}{Brian~M. Sadler},
  \bibinfo{person}{Wei-Lun Chao}, {and} \bibinfo{person}{Yu Su}.}
  \bibinfo{year}{2022}\natexlab{}.
\newblock \bibinfo{title}{LLM-Planner: Few-Shot Grounded Planning for Embodied
  Agents with Large Language Models}.
\newblock
\newblock


\bibitem[\protect\citeauthoryear{Vemprala, Bonatti, Bucker, and
  Kapoor}{Vemprala et~al\mbox{.}}{2023}]%
        {vemprala2023chatgpt}
\bibfield{author}{\bibinfo{person}{Sai Vemprala}, \bibinfo{person}{Rogerio
  Bonatti}, \bibinfo{person}{Arthur Bucker}, {and} \bibinfo{person}{Ashish
  Kapoor}.} \bibinfo{year}{2023}\natexlab{}.
\newblock \showarticletitle{ChatGPT for Robotics: Design Principles and Model
  Abilities}.
\newblock \bibinfo{journal}{\emph{Microsoft Autonomous Systems and Robotics
  Research}}  \bibinfo{volume}{2} (\bibinfo{year}{2023}), \bibinfo{pages}{20}.
\newblock


\bibitem[\protect\citeauthoryear{Wang, Cai, Liu, Ma, and Liang}{Wang
  et~al\mbox{.}}{2023}]%
        {Wang2023DescribeEP}
\bibfield{author}{\bibinfo{person}{Zihao Wang}, \bibinfo{person}{Shaofei Cai},
  \bibinfo{person}{Anji Liu}, \bibinfo{person}{Xiaojian Ma}, {and}
  \bibinfo{person}{Yitao Liang}.} \bibinfo{year}{2023}\natexlab{}.
\newblock \bibinfo{title}{Describe, Explain, Plan and Select: Interactive
  Planning with Large Language Models Enables Open-World Multi-Task Agents}.
\newblock
\newblock


\bibitem[\protect\citeauthoryear{West, Bhagavatula, Hessel, Hwang, Jiang, Bras,
  Lu, Welleck, and Choi}{West et~al\mbox{.}}{2022}]%
        {KnowledgeGraph}
\bibfield{author}{\bibinfo{person}{Peter West}, \bibinfo{person}{Chandra
  Bhagavatula}, \bibinfo{person}{Jack Hessel}, \bibinfo{person}{Jena~D. Hwang},
  \bibinfo{person}{Liwei Jiang}, \bibinfo{person}{Ronan~Le Bras},
  \bibinfo{person}{Ximing Lu}, \bibinfo{person}{Sean Welleck}, {and}
  \bibinfo{person}{Yejin Choi}.} \bibinfo{year}{2022}\natexlab{}.
\newblock \showarticletitle{Symbolic Knowledge Distillation: from General
  Language Models to Commonsense Models}. In
  \bibinfo{booktitle}{\emph{Conference of the North American Chapter of the
  Association for Computational Linguistics: Human Language Technologies}}.
  \bibinfo{publisher}{Association for Computational Linguistics},
  \bibinfo{address}{Seattle, WA, United States}, \bibinfo{pages}{4602--4625}.
\newblock


\bibitem[\protect\citeauthoryear{Yang, Gaglione, Neary, and Topcu}{Yang
  et~al\mbox{.}}{2022}]%
        {Yang2022AutomatonBasedRO}
\bibfield{author}{\bibinfo{person}{Yunhao Yang}, \bibinfo{person}{Jean-Raphael
  Gaglione}, \bibinfo{person}{Cyrus Neary}, {and} \bibinfo{person}{Ufuk
  Topcu}.} \bibinfo{year}{2022}\natexlab{}.
\newblock \bibinfo{title}{Automaton-Based Representations of Task Knowledge
  from Generative Language Models}.
\newblock
\newblock


\end{thebibliography}


\end{document}